\documentclass{article}

\usepackage{microtype}
\usepackage{graphicx}
\usepackage{subfigure}
\usepackage{booktabs} 

\usepackage{hyperref}

\usepackage[accepted]{icml2024}

\usepackage{amsmath}
\usepackage{amssymb}
\usepackage{mathtools}
\usepackage{amsthm}

\usepackage[capitalize,noabbrev]{cleveref}

\usepackage[utf8]{inputenc} 
\usepackage[T1]{fontenc}    
\usepackage{hyperref}       
\usepackage{url}            
\usepackage{booktabs}       
\usepackage{amsfonts}       
\usepackage{nicefrac}       
\usepackage{microtype}      
\usepackage{xcolor}         
\usepackage{natbib}
\usepackage{soul}
\usepackage{graphicx}
\usepackage{changepage}
\usepackage{array}
\usepackage{multirow}
\usepackage{amsfonts}       
\usepackage{makecell}

\theoremstyle{plain}

\theoremstyle{definition}

\theoremstyle{remark}

\usepackage[textsize=tiny]{todonotes}

\begin{document}

\twocolumn[
\icmltitle{Why Would You Suggest That? Human Trust in Language Model Responses}



\icmlsetsymbol{equal}{*}

\begin{icmlauthorlist}
\icmlauthor{Manasi Sharma}{comp}
\icmlauthor{Ho Chit Siu}{comp}
\icmlauthor{Rohan Paleja}{comp}
\icmlauthor{Jaime D. Peña}{comp}
\end{icmlauthorlist}

\icmlaffiliation{comp}{MIT Lincoln Laboratory}

\icmlcorrespondingauthor{Manasi Sharma}{manasi.sharma@ll.mit.edu}
\icmlcorrespondingauthor{Ho Chit Siu}{hochit.siu@ll.mit.edu}
\icmlcorrespondingauthor{Rohan Paleja}{rohan.paleja@ll.mit.edu}
\icmlcorrespondingauthor{Jaime D. Peña}{jdpena@ll.mit.edu}

\icmlkeywords{Machine Learning, ICML}

\vskip 0.3in
]



\printAffiliationsAndNotice{}  

\begin{abstract}
The emergence of Large Language Models (LLMs) has revealed a growing need for human-AI collaboration, especially in creative decision-making scenarios where trust and reliance are paramount. Through human studies and model evaluations on the open-ended News Headline Generation task from the LaMP benchmark, we analyze how the framing and presence of explanations affect user trust and model performance. Overall, we provide evidence that adding an explanation in the model response to justify its reasoning significantly increases self-reported user trust in the model when the user has the opportunity to compare various responses. Position and faithfulness of these explanations are also important factors. However, these gains disappear when users are shown responses independently, suggesting that humans trust all model responses, including deceptive ones, equitably when they are shown in isolation. Our findings urge future research to delve deeper into the nuanced evaluation of trust in human-machine teaming systems. 
\footnote{DISTRIBUTION STATEMENT A. Approved for public release. Distribution is unlimited.

This material is based upon work supported by the Under Secretary of Defense for Research and Engineering under Air Force Contract No. FA8702-15-D-0001. Any opinions, findings, conclusions or recommendations expressed in this material are those of the author(s) and do not necessarily reflect the views of the Under Secretary of Defense for Research and Engineering.}
\end{abstract}

\section{Introduction}
The advent of Large Language Models (LLMs) marks a significant milestone in the realm of artificial intelligence, revolutionizing natural language processing (NLP) by demonstrating remarkable proficiency across diverse tasks like summarization \cite{yang2023exploring} and mathematical reasoning \cite{frieder2023mathematical}, and applicability in domains such as finance \cite{yang2023fingpt} and even robotics \cite{macdonald2024language}. As these models proliferate, we must consider user trust in interactions with LLMs.

Lee and See define trust as ``the attitude that an agent will help achieve an individual’s goals in a situation characterized by uncertainty and vulnerability,'' which must be \textit{calibrated} to avoid overtrust or distrust, the catalysts for overreliance and underuse of systems \cite{lee2004trust}. Mayer et al. argue that \textit{trustworthiness} is affected by the trustee's perceived properties of \textit{ability}, \textit{benevolence}, and \textit{integrity} \cite{mayer1995integrative}.

We target the user personalization setting (LaMP Benchmark \cite{salemi2024lamp} for language model personalization) due to its selection of open-ended text generation tasks and investigate how distinct phrasings of free-form model justifications of responses can affect user trust. We run human studies that gather preferences regarding an array of different model responses, with varying degrees of insight into their decision-making process, and also evaluate the model performance on the task, including these justifications, in order to analyze any tradeoffs that may occur between trust and performance.

We examine the following research questions:\\
\textbf{RQ1} To what extent do the presence and framing of an explanation of a model recommendation impact user trust?\\
\textbf{RQ2} To what extent does the truthfulness of the model explanation of a model recommendation impact user trust?\\
\textbf{RQ3} Are there any tradeoffs between types of explanations that maximize user trust and model performance?

\section{Related Works}
There exists much literature that describes and evaluates the perceived ``trustworthiness'' of language models \cite{sun2024trustllm, wang2024decodingtrust}. However, the literature is largely not founded on actual human studies that evaluate trustworthiness, nor does it typically even define trust. Rather, some studies have used other language models for trust evaluation (sometimes with human audits) \cite{aher2023using}. Moreover, bias and malicious intent are typically used as proxies for trust \cite{sun2024trustllm, wang2024decodingtrust}, but these only examine the benevolence property \cite{mayer1995integrative} of trust, ignoring less nefarious trust-breaking aspects like explanation framing. We choose to focus on trust between the user and the model apart from explicitly biased and harmful behavior. 

Additionally, there are quite a few works that aim to test for or improve logically consistent reasoning (plausibility) \cite{huang2023large, lampinen2022language, marasović2022fewshot, ye2022unreliability, lanham2023measuring, chen2023models} or trust in a human-computer interaction setting \cite{10.1145/3290605.3300717, 10.1145/3301275.3302277}, but they exclusively analyze common-sense reasoning tasks such as QA or classification that involve short number of easily verifiable factual steps. In contrast, we investigate more open-ended creative tasks (such as headline generation) with subjective solutions rather than objective ones to be binarily determined correct or incorrect. Furthermore, some works assess only chain of thought (CoT) explanations or positionality in the form of pre and post-hoc justifications \cite{ye2022unreliability, chen2023models}, but we expand to a wider variety of explanations, including fake justifications and cross-domain thinking. Most significantly, we also evaluate the trustworthiness of model responses through a human study, whereas some other works utilize older, less powerful models that are much less utilized by the general public \cite{bansal2021does} or evaluate the plausibility of the model outputs via heuristic checks \cite{ye2022unreliability, turpin2023language}.

\section{Methodology}
\textbf{Experiment Details} \quad We use model outputs from two state-of-the-art RLHF fine-tuned models, GPT-3.5-Turbo (gpt-3.5-turbo-0613) \cite{ouyang2022training} and GPT-4 \cite{openai2024gpt4} accessed on 04/2024. LaMP (LaMP: When Large Language Models Meet Personalization) \cite{salemi2024lamp} introduces a benchmark for training and evaluating language models to generate personalized outputs. We use the validation set of 1500 samples from the LaMP-4 task (News Headline Generation) for the human study and evaluating model performance [we also evaluate model performance on the LaMP-5 (Research Paper Title Generation), and LaMP-7 (Tweet Paraphrasing) tasks, see \ref{appendix:a}]. To also assess model performance (and determine tradeoffs with respect to trustworthiness), we use the open-ended text generation metric ROUGE score \cite{lin-2004-rouge}.

\begin{table*}
    \centering
    \caption{Explanation styles considered for interacting with the LLM.}
    \footnotesize
    \resizebox{0.8\textwidth}{!}{%
    \begin{tabular}{|l|l|l|l|l|}
    \hline
    \textbf{ID}&\textbf{Explanation Style}&\textbf{\thead{History}}&\textbf{\thead{Justification}}&\textbf{Description}\\
    \hline
    E1&NoHistory&No&None&The model is not required to justify itself and can respond freely.\\
    E2&RetrievedHistory&Yes&None&The model is not required to justify itself and can respond freely.\\
    E3&NoHistoryPrefix&No&Prefix&The model mentions it has thought carefully about its answer as a prefix.\\
    E4&NoHistoryPreJust&No&Pre-hoc&The model first explains its reasoning before responding (like E-P in \cite{ye2022unreliability}).\\
    E5&NoHistoryPostJust&No&Post-hoc&The model first responds and then explains its reasoning. (like P-E in \cite{ye2022unreliability}).\\
    E6&NoHistoryCrossDomainJust&No&Cross-domain&The model reasons on a related topic and draws upon those insights.\\
    E7&NoHistoryFakeJust&No&Fake&The model provides a purposefully false explanation for its response.\\
    E8&RetrievedHistoryPreJust&Yes&Pre-hoc&The model first explains its reasoning before responding.\\
    E9&RetrievedHistoryPostJust&Yes&Post-hoc&The model first responds and then explains its reasoning.\\
    E10&RetrievedHistoryCrossDomainJust&Yes&Cross-domain&The model reasons on a related topic and draws upon those insights.\\
    E11&RetrievedHistoryFakeJust&Yes&Fake&The model provides a purposefully false explanation w.r.t random examples.\\
    \hline
    \end{tabular}
    }
    \label{tab:explanations}
\end{table*}

\textbf{Prompts} \quad We consider 17 different explanation styles for model performance evaluation (see \ref{appendix:a}), but due to survey time constraints, we select 11 for the human study (Table \ref{tab:explanations}). We vary justification type and user history, for which there are two scenarios --- whether or not the model is asked to contextualize its response with respect to some user history related to the prompt. In this context, \textit{user history} is the entire behavior (i.e., text samples) associated with the author (user) of a given text. The model generates zero-shot explanations when it is not provided with the user history. Few-shot explanations are generated when the model is provided with the retrieved history (via RAG), obtained using the top \textit{k=5} most semantically similar previous texts the author generated in the past. Example responses can be found online \footnote{Data and code will be available soon}.

\textbf{Human-AI Trustworthiness Survey} \quad We ran human experiments in the form of a survey to test the self-reported trustworthiness on the News Generation Task in LaMP-4 (as most people consume news sources in some form \cite{pewnews}) by asking them to imagine themselves as a news reporter enlist AI agents for help in finalizing headlines. We then present them with consecutive model responses (with different types of explanations) and ask them to judge them both independently and in relation to each other according to a few criteria discussed below. In-depth set up details are provided in the Appendix (see \ref{appendix:a}). Each response is first independently appraised by the participant on three grounds: 1) \textit{competence} (``This response makes sense.'') 2) \textit{usefulness} (``This response actively helps me decide on a final headline.'') 3) \textit{trust} (``I trust this assistant.''), using a 5-option multiple choice question that translates to Likert scale (1 corresponding to “strongly disagree” and 5 corresponding to “strongly agree.”). After the rating section, three groups of the responses are directly compared with one another, with subjects being asked to rate the responses in each group only in order of trustworthiness. Other work \cite{ouyang2022training, fernandes2023bridging} has demonstrated that ranking data is a more robust representation of human preferences. 

We performed a mixed-effects linear regression on each of the three Likert questions \cite{norman2010likert}, with a categorical independent variable of response type, and predictors of gender, prompt (article snippet), response, age, education, LLM familiarity, computer science familiarity, and AI generated context exposure. Pairwise t-tests were performed on significant factors on the regression. Pairwise Mann-Whitney U tests were performed for the ranking data within each of the three groups. A Bonferroni correction was made for the threshold of statistical significance for pairwise tests.

\section{Results}

\begin{figure*}
    \centering
    \includegraphics[width=0.9\textwidth]{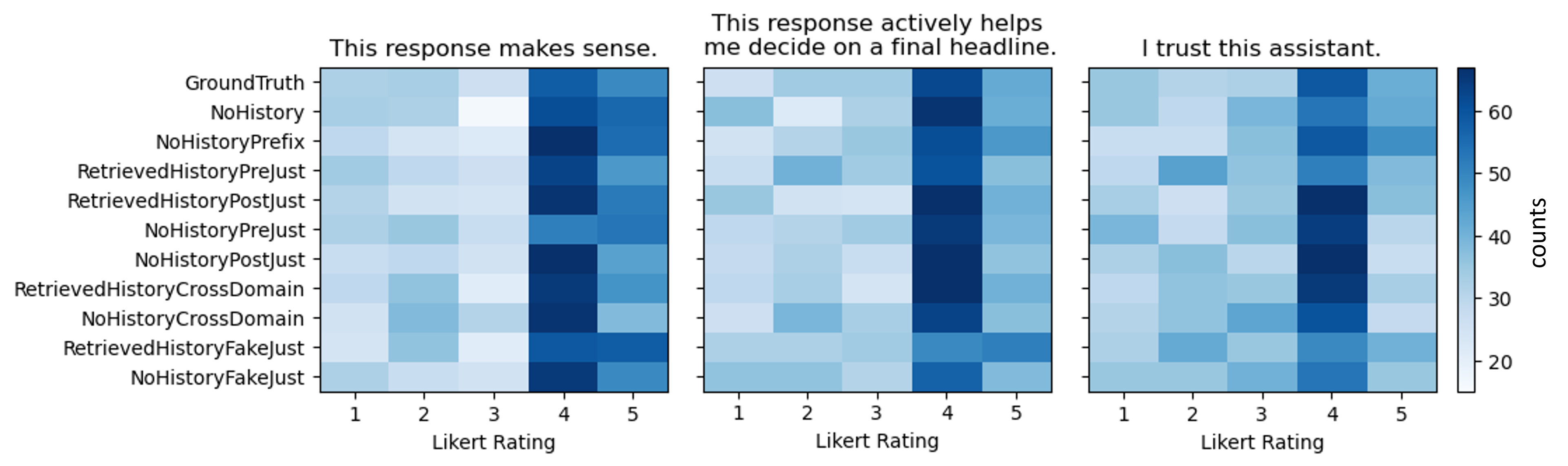}
    \vspace*{-3mm}
    \caption{Combined heat maps and box plots of the three ranking groups. Significance bars omitted for clarity.}
    \label{fig:likert_ratings}
\end{figure*}

\begin{figure}[ht]
    \centering
        \centering
        \includegraphics[width=0.5\textwidth]{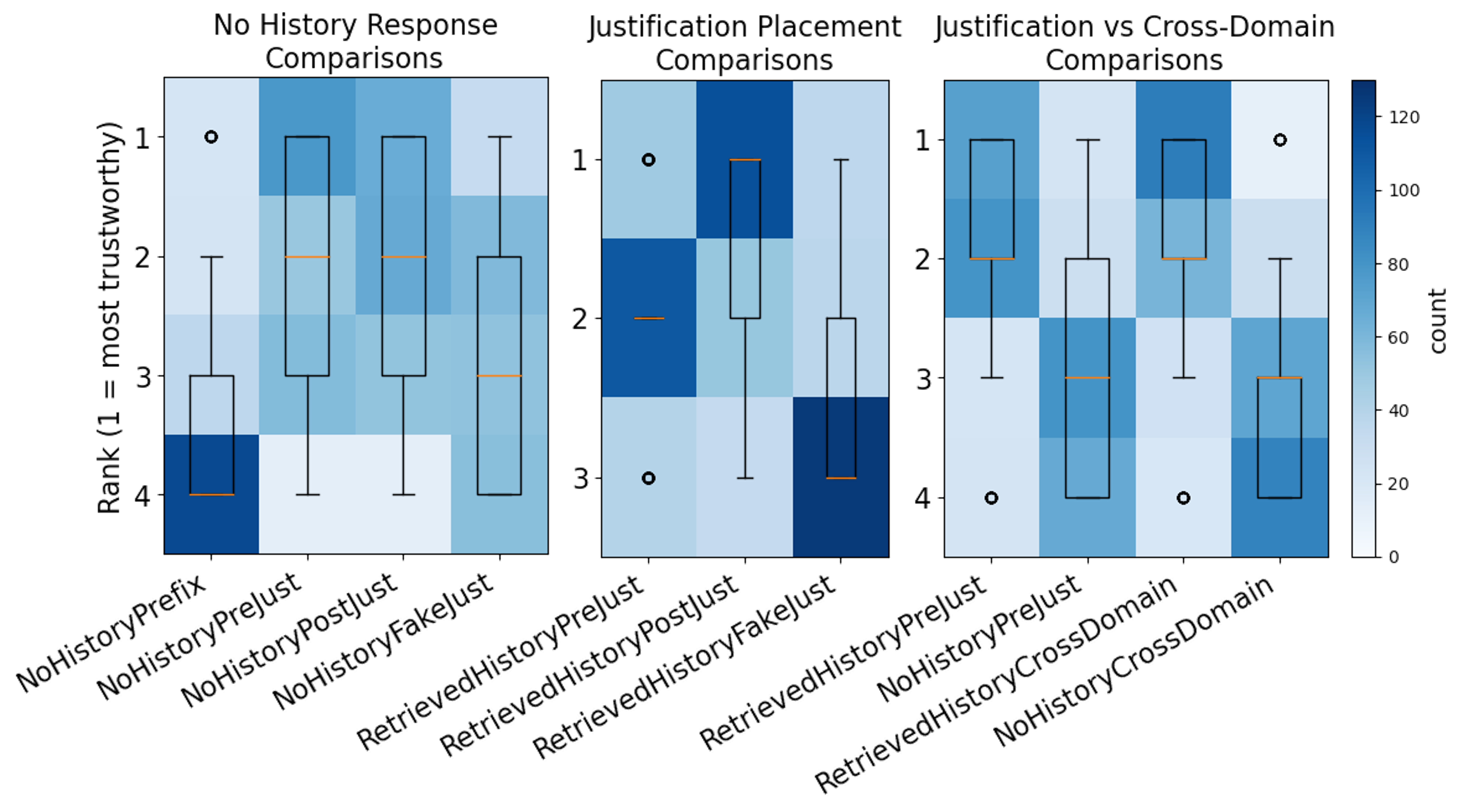} 
        \vspace*{-7mm}
        \caption{Combined heat maps and box plots of the three ranking groups. Significance bars omitted for clarity (see text).}
        \label{fig:fig1}
\vspace*{-0.5mm}
\end{figure}



\textbf{Human Experiment Results} \quad 99 individuals completed the experiment. The Likert ratings for all three questions showed that the majority of people either somewhat or strongly agreed with each statement for all responses (Fig. \ref{fig:likert_ratings}). ``I trust this assistant'' was the only question with somewhat more variation in distribution, but still showed most answers to be Somewhat or Strongly Agree. The regression for each of the questions, with the response type being the independent variable showed no significant differences between the responses generated by the different prompts (\textit{competence} $p=0.86$, \textit{usefulness} $p=0.92$, \textit{trust} $p=0.60$). Self-reported exposure to AI-generated content was found to be a significant factor for \textit{trust} ($p=0.006$), though only the ``occasional'' and the ``very regular'' groups showed significant differences ($p=0.0009$) in post-hoc tests, but with only a small effect size ($d=0.125$). Gender was also found to be a significant factor for all three questions, though pairwise tests showed that it was only in comparisons with groups other than male/female, for which there were only three participants, insufficent for validity. 

Many differences appeared when participants were asked to \textit{rank} response types. The vast majority of the pairwise comparisons of rankings were statistically significantly different from each other (Fig. \ref{fig:fig1}). 
For simplicity, we report the highest statistically-significant $p$ value in each group.
In the No History group, all pairwise comparisons were statistically significantly different ($p\leq6.3\times10^{-8}$) other than the pre- and post-justification conditions ($p=0.53$). 
In the Justification Placement group, all pairwise comparisons were statistically significantly different ($p\leq1.4\times 10^{-7}$).In the Justification vs Cross-Domain group, all pairwise comparisons were significantly different ($p\leq3.5\times 10^{-18}$) other than RetrievedHistoryPrejust and RetrievedHistoryCrossDomain ($p=0.083$) and NoHistoryPrejust and NoHistoryCrossDomain ($p=0.062$).

Subject commentary provided at the end of the experiment indicated a collection of features that were deemed to make the responses less trustworthy, including irrelevant information, hypothesized headlines, lack of explanation, restating the task, the use of metaphors, and stating that the assistant ``thought long and deeply...'' Features that were deemed to make the responses more trustworthy included concise responses, informal language, relevant explanation, lack of explanation, and answer followed by justification. Lack of explanation was cited as \textit{both} a contributor to trustworthiness \textit{and} untrustworthiness, by different people. 


\begin{figure}
        \centering
        \includegraphics[width=0.4\textwidth]{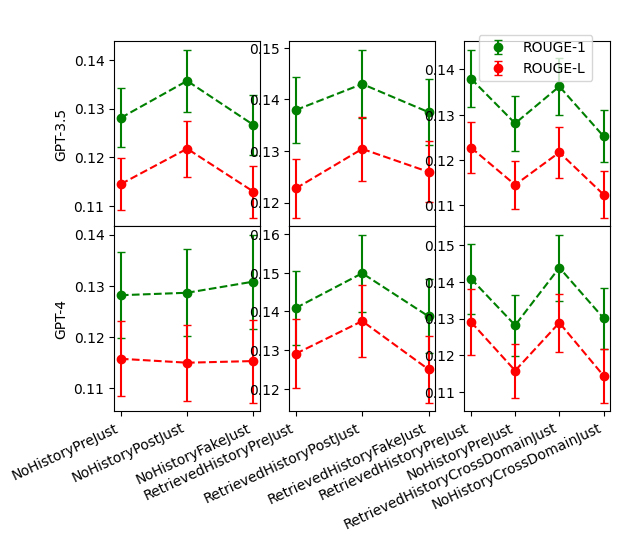} 
        \vspace*{-1mm}
        \caption{ROUGE-1 and ROUGE-L mean and standard error scores for GPT-3.5 and GPT-4 results on LaMP-4 for the comparisons used in the ranking groups.}
        \label{fig:fig2}
\vspace*{-2mm}
\end{figure}

\textbf{Model Performance Evaluation Results} \quad The ROUGE-1 and ROUGE-L scores for GPT-3.5 and GPT-4 results on LaMP-4 for the same comparisons in the ranking experiment (Fig. \ref{fig:fig1}) can be seen in Fig. \ref{fig:fig2}; the error bars are standard error. Overall, the trendlines for model performance (Fig. \ref{fig:fig2}) seem to follow the outlines of trustworthiness ranking results (Fig. \ref{fig:fig1}) (PrefixJust is not considered as it was manually curated). Conversely, due to overlapping error bars, the performance differences do not rise to a statistically significant threshold, with the exception of the RetrievedHistoryPreJust and NoHistoryCrossDomain cases for GPT-3.5. 

However, some interesting observations can still be gleaned from the raw values. Post-hoc justification seems to outpace the Pre-hoc and Fake justifications by a small amount in the No History and Retrieved History cases across GPT-3.5 and GPT-4, while the Pre-hoc, Fake and Cross-domain justifications perform equivalently (see Fig. \ref{fig:fig2} a-b). The explanations based on retrieved data also surpass those without data on model performance. (see Fig. \ref{fig:fig2} c).

\vspace{-2mm}
\section{Discussion}
\vspace{-1mm}

\textbf{Feedback Mechanisms}
We find that human ranking data better elicits statistically significant differences in human preferences (see \ref{fig:fig2}), affirming the basis for its common use in human data collection \cite{ouyang2022training, fernandes2023bridging}. However, in most real-world scenarios, model responses are displayed independently (without side-by-side comparisons), so humans reactions will likely adhere more closely to the Likert ratings (see \ref{fig:likert_ratings}), implying that a difference in trust is not expressed unless asked directly in a zero-sum ranking setting.

\textbf{Adding an Explanation Improves User Trust}
In the ranking setting, we see that presence of an explanation (PreJust, PostJust, FakeJust) improves self-reported user trust over the absence of one (Prefix case) (see \ref{fig:fig2}a). Post-hoc justifications in particular outrank other types of explanations in terms of trust in both the No History and Retrieved History settings, while Pre-hoc and Cross-domain justifications fare similarly. Moreover, displaying a more comprehensive response, through referencing retrieved examples in the justification, also positively impacts user trust (see \ref{fig:fig2}c). However, these increases disappear in the Likert case implying that this preference may be subtle, one the human is not actively aware of. From a practical perspective, this means humans seem to trust any model response (the average score of the Likert being close to 4/5) in the moment it is displayed.

\textbf{The Faithfulness of the Explanation Impacts User Trust}
The willful lack of faithfulness of an explanation (whether the model is asked to actively lie about its reasoning process versus self-described to align with the reasoning of the model) negatively impacts user trust in the both the no history and retrieved history ranking cases. Interestingly however, this result vanishes in the Likert case (see \ref{fig:fig1}) as well, suggesting that humans may identify misleading replies when comparisons are available but are poor at catching deception in an isolated case. This insight has crucial ramifications for human trust in plausibly mis-informative model responses.

\textbf{No Tradeoffs between Explanation Types and Performance}
There is no statistically significant tradeoff between explanation types and model performance, with the exception of the RetrievedHistoryPreJust and NoHistoryCrossDomain cases for GPT-3.5. This potentially demonstrates that, along with the user trust results, no specific type of explanation maximizes user trust and performance. A nuance is that post-hoc justification seems to outperform all other types of explanations (including pre-hoc) across the LaMP-4, 5 and 7 datasets, but it does not rise to a statistical significance.

\textbf{Limitations and Future Work}
A key limitation of our work is the assumption that self-explanations accurately reflect the model's true underlying reasoning process. This notion has been shown to be imprecise in the case of logical questions (eg. \cite{lampinen2022language}) but has not yet been explored extensively for open-ended or creative analyses. A future examination can tackle this - it is possible that the true faithfulness of model justifications could alter the results, although the current insights regarding regarding trust in self-provided explanations are also valuable since, regardless of true accuracy, they may already be used by people prompting for more understanding from the model.
Additionally, our experiment does not exactly follow Lee and See's definition of trust \cite{lee2004trust}, as we do not inculcate situational uncertainty. Furthermore, the original data and the imagined user setting (a news reporter creating a headline) are not personalized to the user, and users are asked to self-report trust instead of it being measured objectively. We are planning studies to better address both deficiencies.

Finally, much of the work in current ML trustworthiness defines trust within the lens of negative impacts such as bias, unfairness, harmful content, misinformation, etc. While such work is vital, longstanding studies in trust \cite{lee2004trust,mayer1995integrative,hoff2015trust} show it to be more inclusive, involving benign aspects such as the response structure and degree of model anthropomorphism. We thus hope to see more work on ML trustworthiness that expands the current cutting-edge and draws on established trust literature in the social sciences.

\vspace{-2mm}
\section{Conclusion}
Humans tend to trust language model responses regardless of the explanation type offered, although when they have comparisons with a variety of responses available, they trust a response with an explanation, especially a post-hoc explanation, more. Importantly, the falsity of explanations is a negative factor for user trust in the ranking setting. Furthermore, as there is no model performance degradation, human-AI trustworthiness would benefit from prompting language models for explanations.

\section{Acknowledgements}
The authors acknowledge MIT Lincoln Laboratory Supercomputing Center for providing the high performance computing resources that have contributed to the research results reported within this paper and the Behavioral Research Lab at MIT for help with recruiting participants. We also acknowledge Jacob Macdonald for his help with feedback and editing, and Vince Mancuso and Kimberly Tanguay for help with clearing the final work.

\bibliographystyle{plainnat}

\bibliography{example_paper}

\begin{thebibliography}{39}
\providecommand{\natexlab}[1]{#1}
\providecommand{\url}[1]{\texttt{#1}}
\expandafter\ifx\csname urlstyle\endcsname\relax
  \providecommand{\doi}[1]{doi: #1}\else
  \providecommand{\doi}{doi: \begingroup \urlstyle{rm}\Url}\fi

\bibitem[Aher et~al.(2023)Aher, Arriaga, and Kalai]{aher2023using}
Gati Aher, Rosa~I. Arriaga, and Adam~Tauman Kalai.
\newblock Using large language models to simulate multiple humans and replicate human subject studies, 2023.

\bibitem[Bansal et~al.(2021)Bansal, Wu, Zhou, Fok, Nushi, Kamar, Ribeiro, and Weld]{bansal2021does}
Gagan Bansal, Tongshuang Wu, Joyce Zhou, Raymond Fok, Besmira Nushi, Ece Kamar, Marco~Tulio Ribeiro, and Daniel~S. Weld.
\newblock Does the whole exceed its parts? the effect of ai explanations on complementary team performance, 2021.

\bibitem[Brown et~al.(2020)Brown, Mann, Ryder, Subbiah, Kaplan, Dhariwal, Neelakantan, Shyam, Sastry, Askell, Agarwal, Herbert-Voss, Krueger, Henighan, Child, Ramesh, Ziegler, Wu, Winter, Hesse, Chen, Sigler, Litwin, Gray, Chess, Clark, Berner, McCandlish, Radford, Sutskever, and Amodei]{brown2020language}
Tom~B. Brown, Benjamin Mann, Nick Ryder, Melanie Subbiah, Jared Kaplan, Prafulla Dhariwal, Arvind Neelakantan, Pranav Shyam, Girish Sastry, Amanda Askell, Sandhini Agarwal, Ariel Herbert-Voss, Gretchen Krueger, Tom Henighan, Rewon Child, Aditya Ramesh, Daniel~M. Ziegler, Jeffrey Wu, Clemens Winter, Christopher Hesse, Mark Chen, Eric Sigler, Mateusz Litwin, Scott Gray, Benjamin Chess, Jack Clark, Christopher Berner, Sam McCandlish, Alec Radford, Ilya Sutskever, and Dario Amodei.
\newblock Language models are few-shot learners, 2020.

\bibitem[Chen et~al.(2023)Chen, Zhong, Ri, Zhao, He, Steinhardt, Yu, and McKeown]{chen2023models}
Yanda Chen, Ruiqi Zhong, Narutatsu Ri, Chen Zhao, He~He, Jacob Steinhardt, Zhou Yu, and Kathleen McKeown.
\newblock Do models explain themselves? counterfactual simulatability of natural language explanations, 2023.

\bibitem[Doshi-Velez and Kim(2017)]{doshivelez2017rigorous}
Finale Doshi-Velez and Been Kim.
\newblock Towards a rigorous science of interpretable machine learning, 2017.

\bibitem[Fernandes et~al.(2023)Fernandes, Madaan, Liu, Farinhas, Martins, Bertsch, de~Souza, Zhou, Wu, Neubig, and Martins]{fernandes2023bridging}
Patrick Fernandes, Aman Madaan, Emmy Liu, António Farinhas, Pedro~Henrique Martins, Amanda Bertsch, José G.~C. de~Souza, Shuyan Zhou, Tongshuang Wu, Graham Neubig, and André F.~T. Martins.
\newblock Bridging the gap: A survey on integrating (human) feedback for natural language generation, 2023.

\bibitem[Frieder et~al.(2023)Frieder, Pinchetti, Chevalier, Griffiths, Salvatori, Lukasiewicz, Petersen, and Berner]{frieder2023mathematical}
Simon Frieder, Luca Pinchetti, Alexis Chevalier, Ryan-Rhys Griffiths, Tommaso Salvatori, Thomas Lukasiewicz, Philipp~Christian Petersen, and Julius Berner.
\newblock Mathematical capabilities of chatgpt, 2023.

\bibitem[Hoff and Bashir(2015)]{hoff2015trust}
Kevin~Anthony Hoff and Masooda Bashir.
\newblock Trust in automation: Integrating empirical evidence on factors that influence trust.
\newblock \emph{Human factors}, 57\penalty0 (3):\penalty0 407--434, 2015.

\bibitem[Huang et~al.(2023)Huang, Mamidanna, Jangam, Zhou, and Gilpin]{huang2023large}
Shiyuan Huang, Siddarth Mamidanna, Shreedhar Jangam, Yilun Zhou, and Leilani~H. Gilpin.
\newblock Can large language models explain themselves? a study of llm-generated self-explanations, 2023.

\bibitem[Jacovi and Goldberg(2020)]{jacovi-goldberg-2020-towards}
Alon Jacovi and Yoav Goldberg.
\newblock Towards faithfully interpretable {NLP} systems: How should we define and evaluate faithfulness?
\newblock \emph{Association for Computational Linguistics}, pages 4198--4205, 2020.
\newblock URL \url{https://aclanthology.org/2020.acl-main.386}.

\bibitem[Kunkel et~al.(2019)Kunkel, Donkers, Michael, Barbu, and Ziegler]{10.1145/3290605.3300717}
Johannes Kunkel, Tim Donkers, Lisa Michael, Catalin-Mihai Barbu, and J\"{u}rgen Ziegler.
\newblock Let me explain: Impact of personal and impersonal explanations on trust in recommender systems.
\newblock In \emph{Proceedings of the 2019 CHI Conference on Human Factors in Computing Systems}, CHI '19, page 1–12, New York, NY, USA, 2019. Association for Computing Machinery.
\newblock ISBN 9781450359702.
\newblock \doi{10.1145/3290605.3300717}.
\newblock URL \url{https://doi.org/10.1145/3290605.3300717}.

\bibitem[Lampinen et~al.(2022)Lampinen, Dasgupta, Chan, Matthewson, Tessler, Creswell, McClelland, Wang, and Hill]{lampinen2022language}
Andrew~K. Lampinen, Ishita Dasgupta, Stephanie C.~Y. Chan, Kory Matthewson, Michael~Henry Tessler, Antonia Creswell, James~L. McClelland, Jane~X. Wang, and Felix Hill.
\newblock Can language models learn from explanations in context?, 2022.

\bibitem[Lanham et~al.(2023)Lanham, Chen, Radhakrishnan, Steiner, Denison, Hernandez, Li, Durmus, Hubinger, Kernion, Lukošiūtė, Nguyen, Cheng, Joseph, Schiefer, Rausch, Larson, McCandlish, Kundu, Kadavath, Yang, Henighan, Maxwell, Telleen-Lawton, Hume, Hatfield-Dodds, Kaplan, Brauner, Bowman, and Perez]{lanham2023measuring}
Tamera Lanham, Anna Chen, Ansh Radhakrishnan, Benoit Steiner, Carson Denison, Danny Hernandez, Dustin Li, Esin Durmus, Evan Hubinger, Jackson Kernion, Kamilė Lukošiūtė, Karina Nguyen, Newton Cheng, Nicholas Joseph, Nicholas Schiefer, Oliver Rausch, Robin Larson, Sam McCandlish, Sandipan Kundu, Saurav Kadavath, Shannon Yang, Thomas Henighan, Timothy Maxwell, Timothy Telleen-Lawton, Tristan Hume, Zac Hatfield-Dodds, Jared Kaplan, Jan Brauner, Samuel~R. Bowman, and Ethan Perez.
\newblock Measuring faithfulness in chain-of-thought reasoning, 2023.

\bibitem[Leahy and Siu(2024)]{leahy2024tell}
Kevin Leahy and Ho~Chit Siu.
\newblock Tell me what you want (what you really, really want): Addressing the expectation gap for goal conveyance from humans to robots.
\newblock \emph{End-User Development for Human-Robot Interaction Workshop at Human Robot Interaction Conference}, 2024.

\bibitem[Lee and See(2004)]{lee2004trust}
John~D Lee and Katrina~A See.
\newblock Trust in automation: Designing for appropriate reliance.
\newblock \emph{Human factors}, 46\penalty0 (1):\penalty0 50--80, 2004.

\bibitem[Lewis et~al.(2020)Lewis, Perez, Piktus, Petroni, Karpukhin, Goyal, K{\"u}ttler, Lewis, Yih, Rockt{\"a}schel, et~al.]{lewis2020retrieval}
Patrick Lewis, Ethan Perez, Aleksandra Piktus, Fabio Petroni, Vladimir Karpukhin, Naman Goyal, Heinrich K{\"u}ttler, Mike Lewis, Wen-tau Yih, Tim Rockt{\"a}schel, et~al.
\newblock Retrieval-augmented generation for knowledge-intensive nlp tasks.
\newblock \emph{Advances in Neural Information Processing Systems}, 33:\penalty0 9459--9474, 2020.

\bibitem[Liedke and Wang(2023)]{pewnews}
Jacob Liedke and Luxuan Wang.
\newblock News platform fact sheet, 2023.
\newblock URL \url{https://www.pewresearch.org/journalism/fact-sheet/news-platform-fact-sheet/}.

\bibitem[Lin(2004)]{lin-2004-rouge}
Chin-Yew Lin.
\newblock {ROUGE}: A package for automatic evaluation of summaries.
\newblock \emph{Association for Computational Linguistics}, pages 74--81, 2004.
\newblock URL \url{https://aclanthology.org/W04-1013}.

\bibitem[Lundberg and Lee(2017)]{lundberg2017unified}
Scott~M Lundberg and Su-In Lee.
\newblock A unified approach to interpreting model predictions.
\newblock \emph{Advances in neural information processing systems}, 30, 2017.

\bibitem[Macdonald et~al.(2024)Macdonald, Mallick, Wollaber, Pe{\~n}a, McNeese, and Siu]{macdonald2024language}
Jacob~P Macdonald, Rohit Mallick, Allan~B Wollaber, Jaime~D Pe{\~n}a, Nathan McNeese, and Ho~Chit Siu.
\newblock Language, camera, autonomy! prompt-engineered robot control for rapidly evolving deployment.
\newblock In \emph{Companion of the 2024 ACM/IEEE International Conference on Human-Robot Interaction}, pages 717--721, 2024.

\bibitem[Malle(2006)]{malle2006mind}
Bertram~F Malle.
\newblock \emph{How the mind explains behavior: Folk explanations, meaning, and social interaction}.
\newblock MIT press, 2006.

\bibitem[Marasović et~al.(2022)Marasović, Beltagy, Downey, and Peters]{marasović2022fewshot}
Ana Marasović, Iz~Beltagy, Doug Downey, and Matthew~E. Peters.
\newblock Few-shot self-rationalization with natural language prompts, 2022.

\bibitem[Mayer et~al.(1995)Mayer, Davis, and Schoorman]{mayer1995integrative}
Roger~C Mayer, James~H Davis, and F~David Schoorman.
\newblock An integrative model of organizational trust.
\newblock \emph{Academy of management review}, 20\penalty0 (3):\penalty0 709--734, 1995.

\bibitem[Norman(2010)]{norman2010likert}
Geoff Norman.
\newblock Likert scales, levels of measurement and the “laws” of statistics.
\newblock \emph{Advances in health sciences education}, 15\penalty0 (5):\penalty0 625--632, 2010.

\bibitem[OpenAI et~al.(2024)OpenAI, Achiam, Adler, Agarwal, Ahmad, Akkaya, Aleman, Almeida, Altenschmidt, Altman, Anadkat, Avila, Babuschkin, Balaji, Balcom, Baltescu, Bao, Bavarian, Belgum, Bello, Berdine, Bernadett-Shapiro, Berner, Bogdonoff, Boiko, Boyd, Brakman, Brockman, Brooks, Brundage, Button, Cai, Campbell, Cann, Carey, Carlson, Carmichael, Chan, Chang, Chantzis, Chen, Chen, Chen, Chen, Chen, Chess, Cho, Chu, Chung, Cummings, Currier, Dai, Decareaux, Degry, Deutsch, Deville, Dhar, Dohan, Dowling, Dunning, Ecoffet, Eleti, Eloundou, Farhi, Fedus, Felix, Fishman, Forte, Fulford, Gao, Georges, Gibson, Goel, Gogineni, Goh, Gontijo-Lopes, Gordon, Grafstein, Gray, Greene, Gross, Gu, Guo, Hallacy, Han, Harris, He, Heaton, Heidecke, Hesse, Hickey, Hickey, Hoeschele, Houghton, Hsu, Hu, Hu, Huizinga, Jain, Jain, Jang, Jiang, Jiang, Jin, Jin, Jomoto, Jonn, Jun, Kaftan, Łukasz Kaiser, Kamali, Kanitscheider, Keskar, Khan, Kilpatrick, Kim, Kim, Kim, Kirchner, Kiros, Knight, Kokotajlo, Łukasz Kondraciuk, Kondrich,
  Konstantinidis, Kosic, Krueger, Kuo, Lampe, Lan, Lee, Leike, Leung, Levy, Li, Lim, Lin, Lin, Litwin, Lopez, Lowe, Lue, Makanju, Malfacini, Manning, Markov, Markovski, Martin, Mayer, Mayne, McGrew, McKinney, McLeavey, McMillan, McNeil, Medina, Mehta, Menick, Metz, Mishchenko, Mishkin, Monaco, Morikawa, Mossing, Mu, Murati, Murk, Mély, Nair, Nakano, Nayak, Neelakantan, Ngo, Noh, Ouyang, O'Keefe, Pachocki, Paino, Palermo, Pantuliano, Parascandolo, Parish, Parparita, Passos, Pavlov, Peng, Perelman, de~Avila Belbute~Peres, Petrov, de~Oliveira~Pinto, Michael, Pokorny, Pokrass, Pong, Powell, Power, Power, Proehl, Puri, Radford, Rae, Ramesh, Raymond, Real, Rimbach, Ross, Rotsted, Roussez, Ryder, Saltarelli, Sanders, Santurkar, Sastry, Schmidt, Schnurr, Schulman, Selsam, Sheppard, Sherbakov, Shieh, Shoker, Shyam, Sidor, Sigler, Simens, Sitkin, Slama, Sohl, Sokolowsky, Song, Staudacher, Such, Summers, Sutskever, Tang, Tezak, Thompson, Tillet, Tootoonchian, Tseng, Tuggle, Turley, Tworek, Uribe, Vallone, Vijayvergiya,
  Voss, Wainwright, Wang, Wang, Wang, Ward, Wei, Weinmann, Welihinda, Welinder, Weng, Weng, Wiethoff, Willner, Winter, Wolrich, Wong, Workman, Wu, Wu, Wu, Xiao, Xu, Yoo, Yu, Yuan, Zaremba, Zellers, Zhang, Zhang, Zhao, Zheng, Zhuang, Zhuk, and Zoph]{openai2024gpt4}
OpenAI, Josh Achiam, Steven Adler, Sandhini Agarwal, Lama Ahmad, Ilge Akkaya, Florencia~Leoni Aleman, Diogo Almeida, Janko Altenschmidt, Sam Altman, Shyamal Anadkat, Red Avila, Igor Babuschkin, Suchir Balaji, Valerie Balcom, Paul Baltescu, Haiming Bao, Mohammad Bavarian, Jeff Belgum, Irwan Bello, Jake Berdine, Gabriel Bernadett-Shapiro, Christopher Berner, Lenny Bogdonoff, Oleg Boiko, Madelaine Boyd, Anna-Luisa Brakman, Greg Brockman, Tim Brooks, Miles Brundage, Kevin Button, Trevor Cai, Rosie Campbell, Andrew Cann, Brittany Carey, Chelsea Carlson, Rory Carmichael, Brooke Chan, Che Chang, Fotis Chantzis, Derek Chen, Sully Chen, Ruby Chen, Jason Chen, Mark Chen, Ben Chess, Chester Cho, Casey Chu, Hyung~Won Chung, Dave Cummings, Jeremiah Currier, Yunxing Dai, Cory Decareaux, Thomas Degry, Noah Deutsch, Damien Deville, Arka Dhar, David Dohan, Steve Dowling, Sheila Dunning, Adrien Ecoffet, Atty Eleti, Tyna Eloundou, David Farhi, Liam Fedus, Niko Felix, Simón~Posada Fishman, Juston Forte, Isabella Fulford, Leo
  Gao, Elie Georges, Christian Gibson, Vik Goel, Tarun Gogineni, Gabriel Goh, Rapha Gontijo-Lopes, Jonathan Gordon, Morgan Grafstein, Scott Gray, Ryan Greene, Joshua Gross, Shixiang~Shane Gu, Yufei Guo, Chris Hallacy, Jesse Han, Jeff Harris, Yuchen He, Mike Heaton, Johannes Heidecke, Chris Hesse, Alan Hickey, Wade Hickey, Peter Hoeschele, Brandon Houghton, Kenny Hsu, Shengli Hu, Xin Hu, Joost Huizinga, Shantanu Jain, Shawn Jain, Joanne Jang, Angela Jiang, Roger Jiang, Haozhun Jin, Denny Jin, Shino Jomoto, Billie Jonn, Heewoo Jun, Tomer Kaftan, Łukasz Kaiser, Ali Kamali, Ingmar Kanitscheider, Nitish~Shirish Keskar, Tabarak Khan, Logan Kilpatrick, Jong~Wook Kim, Christina Kim, Yongjik Kim, Jan~Hendrik Kirchner, Jamie Kiros, Matt Knight, Daniel Kokotajlo, Łukasz Kondraciuk, Andrew Kondrich, Aris Konstantinidis, Kyle Kosic, Gretchen Krueger, Vishal Kuo, Michael Lampe, Ikai Lan, Teddy Lee, Jan Leike, Jade Leung, Daniel Levy, Chak~Ming Li, Rachel Lim, Molly Lin, Stephanie Lin, Mateusz Litwin, Theresa Lopez, Ryan
  Lowe, Patricia Lue, Anna Makanju, Kim Malfacini, Sam Manning, Todor Markov, Yaniv Markovski, Bianca Martin, Katie Mayer, Andrew Mayne, Bob McGrew, Scott~Mayer McKinney, Christine McLeavey, Paul McMillan, Jake McNeil, David Medina, Aalok Mehta, Jacob Menick, Luke Metz, Andrey Mishchenko, Pamela Mishkin, Vinnie Monaco, Evan Morikawa, Daniel Mossing, Tong Mu, Mira Murati, Oleg Murk, David Mély, Ashvin Nair, Reiichiro Nakano, Rajeev Nayak, Arvind Neelakantan, Richard Ngo, Hyeonwoo Noh, Long Ouyang, Cullen O'Keefe, Jakub Pachocki, Alex Paino, Joe Palermo, Ashley Pantuliano, Giambattista Parascandolo, Joel Parish, Emy Parparita, Alex Passos, Mikhail Pavlov, Andrew Peng, Adam Perelman, Filipe de~Avila Belbute~Peres, Michael Petrov, Henrique~Ponde de~Oliveira~Pinto, Michael, Pokorny, Michelle Pokrass, Vitchyr~H. Pong, Tolly Powell, Alethea Power, Boris Power, Elizabeth Proehl, Raul Puri, Alec Radford, Jack Rae, Aditya Ramesh, Cameron Raymond, Francis Real, Kendra Rimbach, Carl Ross, Bob Rotsted, Henri Roussez,
  Nick Ryder, Mario Saltarelli, Ted Sanders, Shibani Santurkar, Girish Sastry, Heather Schmidt, David Schnurr, John Schulman, Daniel Selsam, Kyla Sheppard, Toki Sherbakov, Jessica Shieh, Sarah Shoker, Pranav Shyam, Szymon Sidor, Eric Sigler, Maddie Simens, Jordan Sitkin, Katarina Slama, Ian Sohl, Benjamin Sokolowsky, Yang Song, Natalie Staudacher, Felipe~Petroski Such, Natalie Summers, Ilya Sutskever, Jie Tang, Nikolas Tezak, Madeleine~B. Thompson, Phil Tillet, Amin Tootoonchian, Elizabeth Tseng, Preston Tuggle, Nick Turley, Jerry Tworek, Juan Felipe~Cerón Uribe, Andrea Vallone, Arun Vijayvergiya, Chelsea Voss, Carroll Wainwright, Justin~Jay Wang, Alvin Wang, Ben Wang, Jonathan Ward, Jason Wei, CJ~Weinmann, Akila Welihinda, Peter Welinder, Jiayi Weng, Lilian Weng, Matt Wiethoff, Dave Willner, Clemens Winter, Samuel Wolrich, Hannah Wong, Lauren Workman, Sherwin Wu, Jeff Wu, Michael Wu, Kai Xiao, Tao Xu, Sarah Yoo, Kevin Yu, Qiming Yuan, Wojciech Zaremba, Rowan Zellers, Chong Zhang, Marvin Zhang, Shengjia
  Zhao, Tianhao Zheng, Juntang Zhuang, William Zhuk, and Barret Zoph.
\newblock Gpt-4 technical report, 2024.

\bibitem[Ouyang et~al.(2022)Ouyang, Wu, Jiang, Almeida, Wainwright, Mishkin, Zhang, Agarwal, Slama, Ray, Schulman, Hilton, Kelton, Miller, Simens, Askell, Welinder, Christiano, Leike, and Lowe]{ouyang2022training}
Long Ouyang, Jeff Wu, Xu~Jiang, Diogo Almeida, Carroll~L. Wainwright, Pamela Mishkin, Chong Zhang, Sandhini Agarwal, Katarina Slama, Alex Ray, John Schulman, Jacob Hilton, Fraser Kelton, Luke Miller, Maddie Simens, Amanda Askell, Peter Welinder, Paul Christiano, Jan Leike, and Ryan Lowe.
\newblock Training language models to follow instructions with human feedback, 2022.

\bibitem[Penha and Hauff(2020)]{Penha_2020}
Gustavo Penha and Claudia Hauff.
\newblock What does bert know about books, movies and music? probing bert for conversational recommendation.
\newblock In \emph{Fourteenth ACM Conference on Recommender Systems}, RecSys ’20. ACM, September 2020.
\newblock \doi{10.1145/3383313.3412249}.
\newblock URL \url{http://dx.doi.org/10.1145/3383313.3412249}.

\bibitem[Ribeiro et~al.(2016)Ribeiro, Singh, and Guestrin]{ribeiro2016should}
Marco~Tulio Ribeiro, Sameer Singh, and Carlos Guestrin.
\newblock "why should i trust you?" explaining the predictions of any classifier.
\newblock In \emph{Proceedings of the 22nd ACM SIGKDD international conference on knowledge discovery and data mining}, pages 1135--1144, 2016.

\bibitem[Salemi et~al.(2024)Salemi, Mysore, Bendersky, and Zamani]{salemi2024lamp}
Alireza Salemi, Sheshera Mysore, Michael Bendersky, and Hamed Zamani.
\newblock Lamp: When large language models meet personalization, 2024.

\bibitem[Siu et~al.(2021)Siu, Pe{\~n}a, Chen, Zhou, Lopez, Palko, Chang, and Allen]{siu2021evaluation}
Ho~Chit Siu, Jaime Pe{\~n}a, Edenna Chen, Yutai Zhou, Victor Lopez, Kyle Palko, Kimberlee Chang, and Ross Allen.
\newblock Evaluation of human-ai teams for learned and rule-based agents in hanabi.
\newblock \emph{Advances in Neural Information Processing Systems}, 34:\penalty0 16183--16195, 2021.

\bibitem[Sun et~al.(2024)Sun, Huang, Wang, Wu, Zhang, Li, Gao, Huang, Lyu, Zhang, Li, Liu, Liu, Wang, Zhang, Vidgen, Kailkhura, Xiong, Xiao, Li, Xing, Huang, Liu, Ji, Wang, Zhang, Yao, Kellis, Zitnik, Jiang, Bansal, Zou, Pei, Liu, Gao, Han, Zhao, Tang, Wang, Vanschoren, Mitchell, Shu, Xu, Chang, He, Huang, Backes, Gong, Yu, Chen, Gu, Xu, Ying, Ji, Jana, Chen, Liu, Zhou, Wang, Li, Zhang, Wang, Xie, Chen, Wang, Liu, Ye, Cao, Chen, and Zhao]{sun2024trustllm}
Lichao Sun, Yue Huang, Haoran Wang, Siyuan Wu, Qihui Zhang, Yuan Li, Chujie Gao, Yixin Huang, Wenhan Lyu, Yixuan Zhang, Xiner Li, Zhengliang Liu, Yixin Liu, Yijue Wang, Zhikun Zhang, Bertie Vidgen, Bhavya Kailkhura, Caiming Xiong, Chaowei Xiao, Chunyuan Li, Eric Xing, Furong Huang, Hao Liu, Heng Ji, Hongyi Wang, Huan Zhang, Huaxiu Yao, Manolis Kellis, Marinka Zitnik, Meng Jiang, Mohit Bansal, James Zou, Jian Pei, Jian Liu, Jianfeng Gao, Jiawei Han, Jieyu Zhao, Jiliang Tang, Jindong Wang, Joaquin Vanschoren, John Mitchell, Kai Shu, Kaidi Xu, Kai-Wei Chang, Lifang He, Lifu Huang, Michael Backes, Neil~Zhenqiang Gong, Philip~S. Yu, Pin-Yu Chen, Quanquan Gu, Ran Xu, Rex Ying, Shuiwang Ji, Suman Jana, Tianlong Chen, Tianming Liu, Tianyi Zhou, William Wang, Xiang Li, Xiangliang Zhang, Xiao Wang, Xing Xie, Xun Chen, Xuyu Wang, Yan Liu, Yanfang Ye, Yinzhi Cao, Yong Chen, and Yue Zhao.
\newblock Trustllm: Trustworthiness in large language models, 2024.

\bibitem[Turpin et~al.(2023)Turpin, Michael, Perez, and Bowman]{turpin2023language}
Miles Turpin, Julian Michael, Ethan Perez, and Samuel~R. Bowman.
\newblock Language models don't always say what they think: Unfaithful explanations in chain-of-thought prompting, 2023.

\bibitem[Wang et~al.(2024{\natexlab{a}})Wang, Chen, Pei, Xie, Kang, Zhang, Xu, Xiong, Dutta, Schaeffer, Truong, Arora, Mazeika, Hendrycks, Lin, Cheng, Koyejo, Song, and Li]{wang2024decodingtrust}
Boxin Wang, Weixin Chen, Hengzhi Pei, Chulin Xie, Mintong Kang, Chenhui Zhang, Chejian Xu, Zidi Xiong, Ritik Dutta, Rylan Schaeffer, Sang~T. Truong, Simran Arora, Mantas Mazeika, Dan Hendrycks, Zinan Lin, Yu~Cheng, Sanmi Koyejo, Dawn Song, and Bo~Li.
\newblock Decodingtrust: A comprehensive assessment of trustworthiness in gpt models, 2024{\natexlab{a}}.

\bibitem[Wang et~al.(2024{\natexlab{b}})Wang, Jiang, Chen, Yang, Zhou, Cho, Fan, Huang, Lu, and Yang]{wang2024recmind}
Yancheng Wang, Ziyan Jiang, Zheng Chen, Fan Yang, Yingxue Zhou, Eunah Cho, Xing Fan, Xiaojiang Huang, Yanbin Lu, and Yingzhen Yang.
\newblock Recmind: Large language model powered agent for recommendation, 2024{\natexlab{b}}.

\bibitem[Wei et~al.(2023)Wei, Wang, Schuurmans, Bosma, Ichter, Xia, Chi, Le, and Zhou]{wei2023chainofthought}
Jason Wei, Xuezhi Wang, Dale Schuurmans, Maarten Bosma, Brian Ichter, Fei Xia, Ed~Chi, Quoc Le, and Denny Zhou.
\newblock Chain-of-thought prompting elicits reasoning in large language models, 2023.

\bibitem[Yang et~al.(2023{\natexlab{a}})Yang, Liu, and Wang]{yang2023fingpt}
Hongyang Yang, Xiao-Yang Liu, and Christina~Dan Wang.
\newblock Fingpt: Open-source financial large language models, 2023{\natexlab{a}}.

\bibitem[Yang et~al.(2023{\natexlab{b}})Yang, Li, Zhang, Chen, and Cheng]{yang2023exploring}
Xianjun Yang, Yan Li, Xinlu Zhang, Haifeng Chen, and Wei Cheng.
\newblock Exploring the limits of chatgpt for query or aspect-based text summarization, 2023{\natexlab{b}}.

\bibitem[Ye and Durrett(2022)]{ye2022unreliability}
Xi~Ye and Greg Durrett.
\newblock The unreliability of explanations in few-shot prompting for textual reasoning, 2022.

\bibitem[Yu et~al.(2019)Yu, Berkovsky, Taib, Zhou, and Chen]{10.1145/3301275.3302277}
Kun Yu, Shlomo Berkovsky, Ronnie Taib, Jianlong Zhou, and Fang Chen.
\newblock Do i trust my machine teammate? an investigation from perception to decision.
\newblock In \emph{Proceedings of the 24th International Conference on Intelligent User Interfaces}, IUI '19, page 460–468, New York, NY, USA, 2019. Association for Computing Machinery.
\newblock ISBN 9781450362726.
\newblock \doi{10.1145/3301275.3302277}.
\newblock URL \url{https://doi.org/10.1145/3301275.3302277}.

\end{thebibliography}

\appendix
\section{Appendix}

\subsection{Human-AI Interaction in Creative Settings}

The realization that LLMs generally have a capacity for creativity warrants discussion regarding their incorporation into human decision-making processes. Further, coupling humans with LLMs for decision-making processes holds additional importance, especially while LLMs continue to occasionally hallucinate during inference, as having a human-in-the-loop affords the means to check LLM outputs for clear error. This relationship can be especially advantageous in open-ended settings without a sole definition for correctness. Moreover, the success of this system hinges heavily upon trust and reliance between the human and the model and, thus, the quality of model explanations. The future growth of the use of AI agents as assistants is an important motivator in our selection of the more subjective and open-ended tasks in the LaMP benchmark.

\subsection{Additional Related Works}
\textbf{LLMs and trustworthiness} \quad
Lee and See define trust to be ``the attitude that an agent will help achieve an individual’s goals in a situation characterized by uncertainty and vulnerability,'' and it must be appropriately \textit{calibrated} to avoid overtrust or distrust, which are respectively the catalysts for overreliance and underuse of an AI system \cite{lee2004trust}. Mayer et al. argue that \textit{trustworthiness} is affected by three perceived properties of the trustee: ability, benevolence, and integrity \cite{mayer1995integrative}. Ability and integrity map well to the common AI measures of task performance and predictability, but it is important to note that these factors are \textit{perceived} by the trustor rather than being objective measures.  Interestingly, humans ascribe moral intent or a lack thereof to AI agents, even in fairly abstract settings like card games without communication, and are \textit{more} likely to ascribe malevolence to otherwise high-ability agents \cite{siu2021evaluation}. Agent intent and the forming and breaking of human trust rely heavily on the human's/trustor's expectations of the agent/trustee. LLM collaboration is a particularly interesting form of AI interaction, as LLM --- much more so than other forms of AI --- have language patterns that closely mimic those of entities in the world to which humans have ascribed behavioral and moral agency \cite{malle2006mind}, and at least superficially match the language pragmatics and reasoning characteristics that humans expect of agentic entities \cite{leahy2024tell}.

\textbf{LLMs and interpretability} \quad
Interpretability research is a broad domain that includes model-aware feature attribution explanations like gradient saliancy methods [] and model-agnostic feature attribution explanations such as LIME \cite{ribeiro2016should} and SHAP \cite{lundberg2017unified}. Nonetheless, as we experiment with both open and closed models, and due to the costly nature of running model-agnostic feature attribution explanations when self-explanation methods can perform at par with LIME \cite{ribeiro2016should}, we investigate self-explanations as a method for interpretability. In addition, we cover a wider assortment of explanation styles beyond just pre- and post- hoc explanations.

\textbf{LLMs and recommender systems} \quad
A line of recent work has focused on adapting generative models for service in recommendation tasks - this can be a part of an existing recommendation system pipeline (describing the items to be recommended) \cite{Penha_2020} or to replace the entire pipeline \cite{wang2024recmind}. However, these works principally gauge the models' aptitudes on item recommendation. While this application is very useful for various business use cases, it differs significantly from the kind of preference learning that may happen with regular users when interacting with a language model. Such tasks would fall a great deal more into the more complex category of having longer-form, subjective, and more creative outputs (as opposed to a single item), for example, generating a research paper title; the inputs would also be similarly longer-form (for example, a history of tweets as opposed to quantitative metrics such as click-through rates). We have not been able to find any large, comprehensive benchmarks that specifically tackle more subjective, long-form user preference aside from LaMP \cite{salemi2024lamp}, which is what we used in our experiments.

\subsection{Methodology}
\label{appendix:a}

\textbf{Additional Model Details} \quad We test using two state-of-the-art RLHF fine-tuned models, GPT-3.5-Turbo (gpt-3.5-turbo-0613) and GPT-4 accessed on 04/2024. GPT-3.5 and GPT-4 are both trained using reward models similar to the ones described in \cite{ouyang2022training} for better instruction-following behavior \cite{openai2024gpt4}, but exact information about parameter count, architecture, and training data has not been disclosed publicly. We employ the full evaluation sets for both models and set the temperature for both models to 0 to control for any differences that may arise from distinct decoding algorithms as opposed to dissimilar prompts and requests for justifications. The OpenAI-API was used and one dataset took around 10 hours to evaluate on one model across all evaluation types.

\textbf{Additional Data Details} \quad LaMP (LaMP: When Large Language Models Meet Personalization) \cite{salemi2024lamp} introduces a novel benchmark aimed at training and evaluating language models to generate personalized outputs. LaMP provides a thorough assessment framework including various language tasks and comprehensive user behavior history encapsulated in profiles. These profiles are either split across users (user-based) or for a single user (time-based). We utilize the time-based versions of the data to zero in on learning preferences for a single user over time, which is a more practical scenario for the use AI agent assistants. We utilize the Validation Sets for each of the tasks (LaMP-4: News Headline Generation, LaMP-5: Research Paper Title Generation, and LaMP-7: Tweet Paraphrasing), which comprise of 1500, 1500 and 1498 samples respectively. This leaves us with 4498 samples for the evaluation. For the GPT-4 model evaluation, 300 examples are randomly selected due to cost constraints.

\textbf{Additional Task Details} \quad The LaMP benchmark comprises seven personalized tasks, covering three text classification tasks and four text generation tasks. We focus on the more complex text generation tasks, as opposed to classification, and select three of the tasks for experimentation (the fourth, Email Subject Generation, is not open-source):\\
\textbf{T1} News Headline Generation: the model is provided with an article snippet usually 1-2 sentences long (assumed to be the first line from the article) and asked to extrapolate and base a headline on it.\\
\textbf{T2} Research Paper Title Generation: the model is similarly provided with an abstract and has to predict the research paper's title.\\
\textbf{T3} Tweet Paraphrasing: the model has to convert a description of a tweet to a unique tweet in the user's unique style.

\textbf{Additional Prompt Details}
\textbf{Prompts} \quad The full list of 17 explanation styles is summarized in Table \ref{tab:explanations}.

We first require five primary different types of justifications the model should consider with its responses, passed into the prompt. 

\begin{itemize}
  \item \textit{None}: the model is not required to justify itself and can respond freely.
  \item \textit{Prefix}: the model is not queried differently from the "None" case, but a prefix of the phrase "After thinking deeply and considering a wide range of possibilities, I came to the conclusion that the most appropriate headline is" is manually pre-pended to the model response.
  \item \textit{Pre-hoc}: the model is first queried to provide a self-explanation of its reasoning before responding. This is similar to the E-P (explain-then-predict) paradigm in \cite{ye2022unreliability}.
  \item \textit{Post-hoc}: query the model for a response first, followed by its self-explanation. This is similar to the P-E (predict-then-explain) paradigm in \cite{ye2022unreliability}.
  \item \textit{Cross-domain hypothesization}: elicit the model's adjacent reasoning capabilities by asking it to first reason in a different but similar domain and then using the insights drawn.
  \item \textit{Fake}: queries the model to provide a completely false explanation to the user which differed from its true reasoning process. In the retrieved case, the model provides a purposefully false explanation for its response with regards to randomly retrieved examples (instead of examples retrieved due to similarity).
\end{itemize}

The model's baseline response is given when no justification is required. \textit{Prefix} explanations test how comfortable the user is with the model claiming to have performed some reasoning, without that reasoning actually being provided. Since this type of response is manually curated, it \textbf{only appears in the human study and not in the model evaluation}. We also only assessed it in the no history case. \textit{Pre-hoc} and \textit{post-hoc} justifications provide direct rationale of the model's response (Note: some works hypothesize that if the generated explanation is post-hoc, the explanation has more ways to be unfaithful. \cite{lanham2023measuring}). \textit{Cross-domain} hypothesization provides insight into the model's generalization capabilities and how such second-order levels of cognition impact model performance (i.e., change in the model's outputs when related but distinct domain information is provided), and whether human users might find them well-founded or extraneous. For example, in the news headline generation task, postulating what a given user, with headlines written in Politics, may write in another area like Economics. Lastly, \textit{fake} justifications probe the aspect of faithfulness by testing whether providing \textit{any} explanation, regardless of the fidelity to the model's self-described true reasoning process, is sufficient for the sustenance of human reliance.

Finally, to understand whether engendering any types of explanations in model responses affects its ultimate performance, we include the following baselines:
\begin{itemize}
    \item \textit{Zero-shot Chain-of-thought (CoT)}: Based on \cite{wei2023chainofthought}, the model is asked to think step-by-step before answering. We consider both the retrieved case and no history case.
    \item \textit{Few-shot Chain-of-thought (CoT)}: Based on \cite{lewis2020retrieval, brown2020language}, the model is asked to think step-by-step by following the reasoning given in examples. Only the No History case is considered, as the examples were fixed (not retrieved) due to resource constraints, and the reasoning for a preselected set of examples was written by a member of the lab.
    \item \textit{Numerical}: Extending the approach of token approach in \cite{huang2023large}, the model is asked to provide numerical values indicating importance to each of its retrieved examples (as opposed to individual tokens). Thus, only the retrieved case is considered.
    \item \textit{Entire History}: Model evaluation given the largest chunk possible of the user history (k=25).
    \item \textit{Random History}: Randomly selected examples not related to the prompt are provided (k=5).
\end{itemize}

In order to perform the ROUGE score analyses of the headlines, we extract the headlines from the explanations using RegEx from a fixed output prompt template the model is prompted with.

\begin{table*}
    \centering
    \caption{Full list of explanation styles considered for model evaluation.}
    \footnotesize
    \resizebox{0.8\textwidth}{!}{%
    \begin{tabular}{|l|l|l|l|l|}
    \hline
    \textbf{ID}&\textbf{Explanation Style}&\textbf{\thead{History}}&\textbf{\thead{Justification}}&\textbf{Description}\\
    \hline
    E1&NoHistory&No&None&The model is not required to justify itself and can respond freely.\\
    E2&RetrievedHistory&Yes&None&The model is not required to justify itself and can respond freely.\\
    E3&RandomHistory&Yes&None&The model is not required to justify itself and can respond freely.\\
    E4&EntireHistory&Yes&None&The model is not required to justify itself and can respond freely.\\
    E5&NoHistoryPrefix&No&Prefix&The model mentions it has thought carefully about its answer as a prefix.\\
    E6&NoHistoryPreJust&No&Pre-hoc&The model first explains its reasoning before responding.\\
    E7&NoHistoryPostJust&No&Post-hoc&The model first responds and then explains its reasoning.\\
    E8&NoHistoryCrossDomainJust&No&Cross-domain&The model reasons on a related topic and draws upon those insights.\\
    E9&NoHistoryFakeJust&No&Fake&The model provides a purposefully false explanation for its response.\\
    E10&NoHistoryZeroShotCoT&No&Zero-Shot&The model is asked to think step-by-step before answering. \cite{wei2023chainofthought}\\
    E11&NoHistoryFewShotCoT&No&Few-Shot&The model follows the step-by-step reasoning given in examples. \cite{lewis2020retrieval, brown2020language}\\
    E12&RetrievedHistoryPreJust&Yes&Pre-hoc&The model first explains its reasoning before responding.\\
    E13&RetrievedHistoryPostJust&Yes&Post-hoc&The model first responds and then explains its reasoning.\\
    E14&RetrievedHistoryCrossDomainJust&Yes&Cross-domain&The model reasons on a related topic and draws upon those insights.\\
    E15&RetrievedHistoryFakeJust&Yes&Fake&The model provides a purposefully false explanation for its response.\\
    E16&RetrievedHistoryZeroShotCoT&Yes&Zero-Shot&The model is asked to think step-by-step before answering. \cite{wei2023chainofthought}\\
    E17&RetrievedHistoryNumericalJust&Yes&Numerical&The model provides numbers indicating the importance each example. \cite{huang2023large}\\
    \hline
    \end{tabular}
    }
    \label{tab:explanations_full}
\end{table*}

Table \ref{tab:prompts_and_responses} displays sample prompts supplied to the model for the different explanation styles.

Many studies discover a trade-off between performance and interpretability (Narang et al. 2020;
Subramanian et al. 2020; Hase et al. 2020, i.a.), while others observe a positive impact
on the performance from including explanations (e.g., CoT-style prompting).

\begin{table*}
    \centering
    \caption{List of example prompt templates and corresponding models responses. The same context and initial prompt are provided to the model for each of the explanation styles: \\ \{context\}: You are a talented journalist who excels at writing headlines.\\ \{prompt\} (this is a specific example): "Generate a headline for the following article snippet: Being an ex wife was very unexpected. I went into it kicking and screaming.  And drunk texting.  Oh-and a little bit of stalking.  To those going through it now I can tell you, you will survive."}
    \footnotesize
    \resizebox{1.0\textwidth}{!}{%
    \begin{tabular}{|l|l|l|l|}
    \hline
    \textbf{ID}&\textbf{Explanation Style}&\textbf{Prompt}&\textbf{Notes}\\
    \hline
    E1&NoHistory&\thead{\{context\}\\ \{prompt\}}&\\
    \midrule
    E2&RetrievedHistory&\thead{\{context\}\\ \{prompt\}\\ Here is the history of articles and their corresponding headlines:\\ \{retrieved\_{data}\}\\}& 
    \thead{\{retrieved\_{data}\} is the k=5 most similar retrieved \\ examples from the user history.}\\
    \midrule
    E3&RandomHistory&\thead{\{context\}\\ \{prompt\}\\ Here is the history of articles and their corresponding headlines:\\ \{random\_{retrieved\_{data}}\}\\}& 
    \thead{\{random\_{retrieved\_{data}}\} is k=5 randomly selected \\ examples from the user history.}\\
    \midrule
    E4&EntireHistory&\thead{\{context\}\\ \{prompt\}\\ Here is the history of articles and their corresponding headlines:\\ \{entire\_{retrieved\_{data}}\}\\}&
    \thead{\{entire\_{retrieved\_{data}}\} is as much of the user history \\ that can fit into context length (which ends up being k=25 \\ of the most similarly retrieved examples)}\\
    \midrule
    E5&NoHistoryPrefix&\thead{\{context\}\\ \{prompt\}}&\\
    \midrule
    E6&NoHistoryPreJust&
    \thead{\{context\}\\ \{prompt\}\\ Please first provide an explanation for how you chose to generate the \\ headline and then the generated headline with the format "Generated Headline: "}
    & \\
    \midrule
    E7&NoHistoryPostJust&
    \thead{\{context\}\\ \{prompt\}\\ Please provide first the generated headline with the format "Generated  \\ Headline: " and then an explanation for how you chose to generate the headline.}
    &\\
    \midrule
    E8&NoHistoryCrossDomainJust&
    \thead{\{context\}\\ \{prompt\}\\ First, please hypothesize 5 examples of what other headlines the person who wrote these headlines would write in another \\ domain, like Art, Science, Politics or Pop Culture. Then, based on the insights  you gained from reasoning on the hypothesized headlines, generate the final answer \\ (the headline for the initial article)  with the format "Generated Headline: ".}
    &\\
    \midrule
    E9&NoHistoryFakeJust&
    \thead{\{context\}\\ \{prompt\}\\ Please provide first the generated headline with the format "Generated \\ Headline: " and then provide a completely fake explanation for how you chose to generate the headline \\ (that is different from the actual logic you followed to reach the answer).}
    &\\
    \midrule
    E10&NoHistoryZeroShotCoT&
    \thead{\{context\}\\ \{prompt\}\\ Please provide the generated headline with the format "Generated Headline: "\\ Let's think step by step. }
    &\\
    \midrule
    E11&NoHistoryFewShotCoT&
    \thead{\{context\}\\ \{prompt\}\\ Here are examples of how headlines are constructed from their corresponding articles:\\
    Article text: "Above all, show your love. Show up. Say something. Do something. Be willing to stand beside the \\ gaping hole that has opened in your friend's life, without flinching or turning away. Be love. Love is the thing that lasts."\\
    Explanation: With the hint that a "gaping hole" has opened up in your friend's life and the recommendation \\ that you should be there for your friend, we can glean that the friend must be experience the grief of a loss.\\
    Headline: "How to Help a Grieving Friend: 11 Things to Do When You're Not Sure What to Do"\\
    ...
    }
    &
    \thead{The examples used for reasoning are fixed (not \\
    retrieved) due to resource constraints in labeling \\ reasoning for all the samples in the val. set.}\\
    \midrule
    E12&RetrievedHistoryPreJust&
    \thead{\{context\}\\ \{prompt\}\\ Here is the history of articles and their corresponding headlines:\\ \{retrieved\_{data}\}\\ Please first provide an explanation for how you chose to generate \\ the headline with respect to the history of previous headlines and their articles, and \\ then the generated headline with the format "Generated Headline: "}
    &\\
    \midrule
    E13&RetrievedHistoryPostJust&
    \thead{\{context\}\\ \{prompt\}\\ Here is the history of articles and their corresponding headlines:\\ \{retrieved\_{data}\}\\ Please provide first the generated headline with the format "Generated \\ Headline: " and then an explanation for how you chose to generate \\ the headline with respect to the history of  previous headlines and their articles.}
    &\\
    \midrule
    E14&RetrievedHistoryCrossDomainJust&
    \thead{\{context\}\\ \{prompt\}\\ Here is the history of articles and their corresponding headlines: \{retrieved\_{data}\}. First, in addition to considering the history of previous headlines, \\ please hypothesize 5 examples of what other headlines the person who wrote these headlines \\ would write in another domain, like Art, Science, Politics or Pop Culture. Then, based on the history of articles and their corresponding headlines and the \\ insights you gained from reasoning on the hypothesized headlines, generate the final answer (the headline for the initial article) with the format  "Generated Headline: ". }
    &\\
    \midrule
    E15&RetrievedHistoryFakeJust&
    \thead{\{context\}\\ \{prompt\}\\ Here is a fake history of unrelated articles and their corresponding headlines:\\ \{fake\_{retrieved\_{data}}\} Now, provide an fake explanation for how you chose to generate the headline with direct \\ reference to the fake history of the given previous headlines (that is different from the actual logic you  followed to reach the answer).}
    &\thead{The fake explanations in this case are with \\ respect to randomly selected examples.\\ .}\\
    \midrule
    E16&RetrievedHistoryZeroShotCoT&
    \thead{\{context\}\\ \{prompt\}\\ Here is the history of articles and their corresponding headlines:\\ \{retrieved\_{data}\}\\ Let's think step by step.}
    &\\
    \midrule
    E17&RetrievedHistoryNumericalJust&
    \thead{\{context\}\\ \{prompt\}\\ Here is the history of articles and their corresponding headlines: \{retrieved\_{data}\} Please first \\ provide the generated headline with the format "Generated Headline: " Then, you must analyze the importance of each example \\ in contributing to your final generated headline in the Python tuple format: (<example>, <float importance>). The importance should be a decimal \\ number to two decimal places ranging from 0 to 1, with 0 implying no contribution at all sentiment and 1 \\ implying the highest contribution. All the float importances should sum to 1.}
    &\\
    \hline
    \end{tabular}
    }
    \label{tab:prompts_and_responses}
\end{table*}

\textbf{Survey Setup Details} \quad We ran human experiments in the form of a survey to test the self-reported trustworthiness that humans experience based on various types of explanations. We operate within the same LaMP dataset \cite{salemi2024lamp} as the model performance evaluation experiments and use the outputs from GPT-3.5 model to display to the subjects. GPT-3.5 was used instead of GPT-4 as more people would have interacted with the GPT-3.5 model due to its proximity to ChatGPT (as opposed to the GPT-4 which requires a paywall). In order to probe a wider variety of the general public, we chose to exclusively employ the News Generation Task in LaMP, as most people consume news sources in some form \cite{pewnews}, while research paper title generation and even the tweet paraphrasing task might require either specialized knowledge of a niche topic or unique features like slang. This allows us to ask subjects to put themselves in a scenario where they can see themselves interacting with an AI agent, i.e. as a news reporter enlist AI agents for help in finalizing headlines. 

We then present them with consecutive explanations and ask them to judge them both independently and in relation to each other according to a few criteria discussed below. In order to balance our desire to display a variety of outputs to subjects with the need to receive adequate signal for each specific output to overcome noise, out of the 1500 samples in the development set, we randomly selected 16 examples of article snippets and their corresponding headlines (with a seed of 23) for use in the human studies. We also pruned and resampled examples that dealt with confounding topics that may adversely impact trustworthiness of responses such as politics until we were left with 16 un-confounded examples.

The survey flow was designed as follows: participants were first introduced to the experiment and asked to envision themselves as star news reporters that must enlist the help of assistants, whose responses they will be evaluating for credibility. The instruction was then followed with a brief familiarization game of matching news headlines to their topic areas to check for participant attention. If they were able to pass this stage, the subjects would proceed to the actual experiment. They would be shown a task instruction: to generate a news headline based on a subsequent article snippet. Then the various responses from different prompts to the model (from the model evaluation section), which encapsulate the discrete explanation styles, would be exhibited to the subject one-at-a-time to assess according to a few criteria. To enable this, one article snippet is randomly selected from the 16 examples (seed set to 23). 

Each response is first independently appraised one three grounds: 1) \textit{competence}: ``This response makes sense'' 2) \textit{usefulness}: ``This response actively helps me decide on a final headline'' and 3) \textit{trust} ``I trust this assistant'', using a 5-option multiple choice question that translates to Likert scale (1 corresponding to “strongly disagree” and 5 corresponding to “strongly agree.”). After all the responses are displayed, a few subsets of the responses are directly compared with one another through ranking, as some other work \cite{ouyang2022training, fernandes2023bridging} has demonstrated that ranking data is a more robust representation of human preferences. We compared:
\begin{enumerate}
    \item Prefix, PreJust, PostJust and FakeJust in the NoHistory case
    \item PreJust, PostJust and FakeJust in the RetrievedHistory case
    \item PreJust and CrossDomainJust both in the cases of NoHistory and RetrievedHistory to compare with the CrossDomain explanation type as well
\end{enumerate}
We were able to cover most of the explanations at least once in the ranking data through these comparisons.

We sought to design the survey with a timeframe of 15 minutes in mind to maximize the number of participants who could participate (eg. taking time out on a coffee or lunch break). Displaying all 14 responses from the model evaluation section to the user would exceed that time limit due to the increased number of Likert scale evaluations (3 per additional response) and the added time and mental overhead in ranking more than 4 responses. Therefore, we decided to limit our responses to ensure that condition, and entire experiment was timed in the pilot phase to be approximately 15 minutes, though actual experiment time tended to be somewhat longer. 

To cut down, we elected to remove "RetrievedHistory" and "EntireHistory" responses, as the responses contain no justification and may not look meaningfully separate to a human from the "NoHistory" response. Moreover, we decided to eliminate both of the CoT questions as the zero-shot case is stylistically very similar to the pre-hoc justification method of explanation-then-answer and the few-shot case follows human-logic that is externally applied and may not be true to how the explanation is actually generated. Finally, we felt that the attribution question stood a bit alone and could not be compared across the no history and retrieved history conditions, and thus though we felt it was a very compelling and distinctive type of explanation, we determined to push it for a future experiment.

This results in 9 remaining responses. We were also curious whether simply mentioning in the response that the model had pondered the prompt and considered many possibilities, regardless of the truth of that claim, could affect trust; subsequently, we added a model response that simply prepended such a prefix before stating the generated headline. Additionally, to account for a potential blanket bias by a participant against any AI generated content, which may impact measures of trustworthiness, we add in the ground truth human response (which is the headline written by the original journalist who wrote the article snippet). We also make it clear to participants that responses may be AI-model generated or generated by a human. This results in a total of 11 responses displayed to the user. After the Likert questions, participants were asked to directly rank the trustworthiness of responses in three groups (Table \ref{fig:fig1}).

Then, this entire procedure of Likert ratings and ranking is repeated with another randomly selected article snippet. Essentially, the experiment has 2 rounds, with 2 article snippets, 11 responses, and 3 ranking groups each. Therefore, each person is answering 66 Likert questions in total, and for each type of explanation. The order of displaying the responses for both Likert and ranking was randomized to prevent biases arising from the sequence of responses.

Participants received a \$5 USD gift card at the end of their experiment. Each experiment took approximately 30 minutes. The total amount spent on participant payment was \$500. Experiments were conducted virtually and asynchronously through supplying online surveys links hosted on Qualtrics.

\subsection{Additional LLM Results}
\label{appendix:b} 
\begin{figure}[!htb]
\minipage{0.4\textwidth}
  \includegraphics[width=\linewidth]{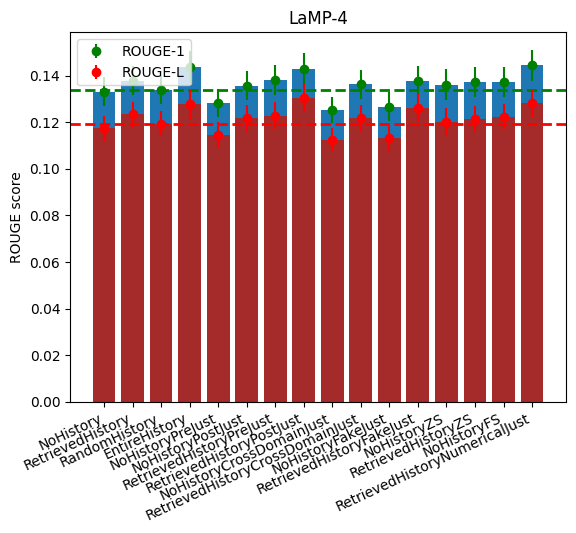}
\endminipage\hfill
\minipage{0.4\textwidth}
  \includegraphics[width=\linewidth]{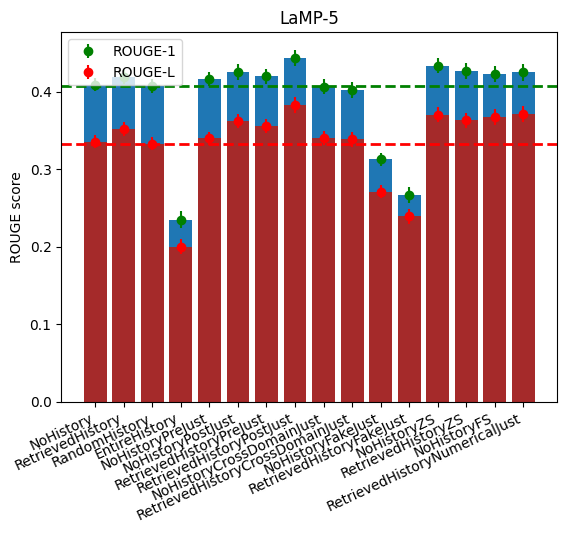}
\endminipage\hfill
\minipage{0.4\textwidth}%
  \includegraphics[width=\linewidth]{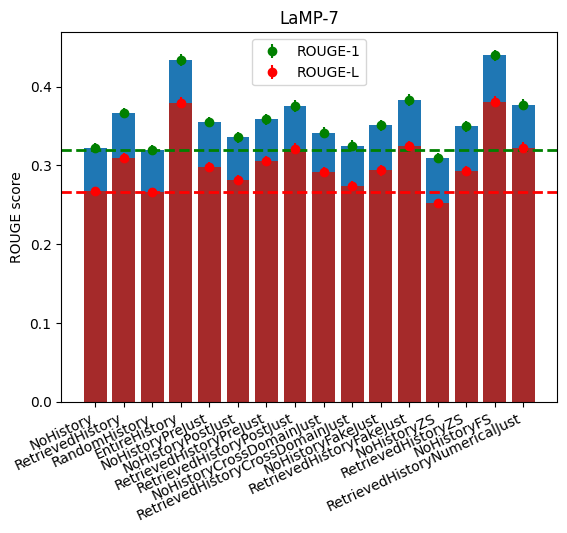}
\caption{Bar Graph visualization of the results in Table \ref{table:full_results_gpt3} for LaMP-4, LaMP-5 and LaMP-7 for GPT-3.5. The random history baseline is visualized as a dotted line to compare other results to.}
\label{fig:lamp}
\endminipage
\end{figure}

\begin{figure}[!htb]
\minipage{0.4\textwidth}
  \includegraphics[width=\linewidth]{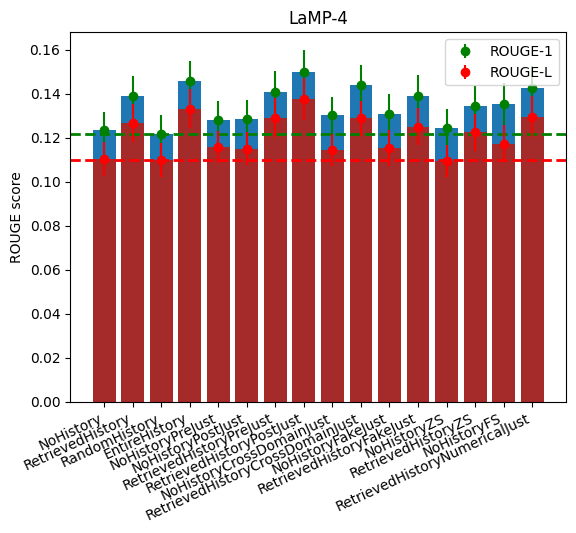}
\endminipage\hfill
\minipage{0.4\textwidth}
  \includegraphics[width=\linewidth]{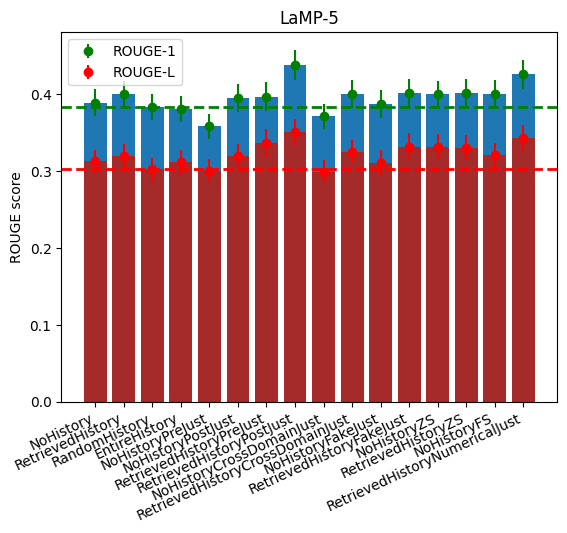}
\endminipage\hfill
\minipage{0.4\textwidth}%
  \includegraphics[width=\linewidth]{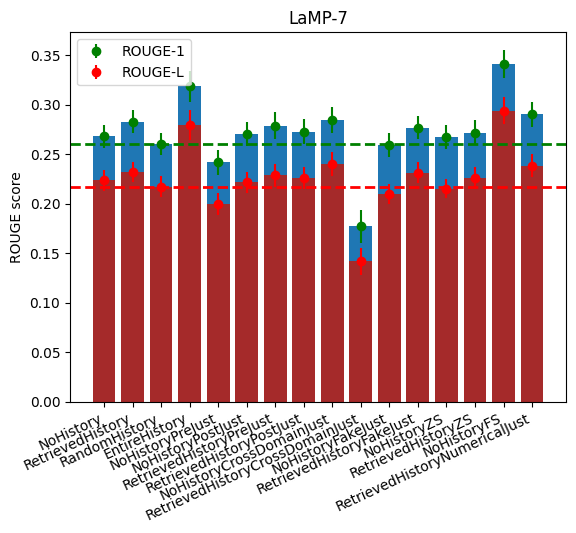}
\caption{Bar Graph visualization of the results in Table \ref{table:full_results_gpt4} for LaMP-4, LaMP-5 and LaMP-7 for GPT-4. The random history baseline is visualized as a dotted line to compare other results to.}
\label{fig:lamp_gpt4}
\endminipage
\end{figure}

The ROUGE-1 and ROUGE-L results for every model response style in Tables \ref{table:full_results_gpt3} and \ref{table:full_results_gpt4} across GPT-3.5 and GPT-4 for LaMP-4, LaMP-5, LaMP-7, along with corresponding standard error bars. These values are also visualized in bar plots in Figures \ref{fig:lamp} and \ref{fig:lamp_gpt4}.

\begin{table*}
   \caption{The mean ROUGE-1 and ROUGE-L scores for each method, tested on GPT-3.5, with accompanying one-sigma standard error bars. The highest ROUGE-1 and ROUGE-L scores in each subsection (No History and Retrieved History) for each dataset are highlighted in blue, while the lowest scores are highlighted in red. The "Other" section (Entire and Random History) is highlighted if the scores are higher than in all the other sections.}
   \centering
\begin{tabular}{|c|c|c|ccc|}
    \hline
    \multirow{2}{*}{} & \multirow{2}{*}{} & \multirow{2}{*}{} & \multicolumn{3}{c}{GPT-3.5} \\ \midrule 
    Type & Method & Metric & LaMP-4 & LaMP-5 & LaMP-7 \\ \midrule
    \multirow{14}{*}{No History}        & Plain           & ROUGE-1 ($\uparrow$) & 0.133 $\pm$ 0.00626 & 0.409 $\pm$ 0.00877 & 0.322 $\pm$ 0.00610  \\
                    &                 & ROUGE-L ($\uparrow$) & 0.117 $\pm$ 0.00544 & 0.335 $\pm$ 0.00814 & 0.267 $\pm$ 0.00561 \\
                    \cmidrule{2-6}
                    & PreJust         & ROUGE-1 ($\uparrow$) & 0.128 $\pm$ 0.00605 & 0.416 $\pm$ 0.00890  & 0.355 $\pm$ 0.00676 \\
                    &                 & ROUGE-L ($\uparrow$) & 0.115 $\pm$ 0.00535 & 0.340 $\pm$ 0.00837  & 0.298 $\pm$ 0.00621 \\
                    \cmidrule{2-6}
                    & PostJust        & ROUGE-1 ($\uparrow$) & \textcolor{blue}{\textbf{0.136}} $\pm$ 0.00638 & 0.425 $\pm$ 0.00986 & 0.336 $\pm$ 0.00670  \\
                    &                 & ROUGE-L ($\uparrow$) & \textcolor{blue}{\textbf{0.122}} $\pm$ 0.00575 & 0.362 $\pm$ 0.00940  & 0.281 $\pm$ 0.00613 \\
                    \cmidrule{2-6}
                    & CrossDomainJust & ROUGE-1 ($\uparrow$) & \textcolor{red}{0.125} $\pm$ 0.00570  & 0.406 $\pm$ 0.00950  & 0.342 $\pm$ 0.00679 \\
                    &                 & ROUGE-L ($\uparrow$) & \textcolor{red}{0.112} $\pm$ 0.00510  & 0.340 $\pm$ 0.00902  & 0.292 $\pm$ 0.00635 \\
                    \cmidrule{2-6}
                    & FakeJust        & ROUGE-1 ($\uparrow$) & 0.127 $\pm$ 0.00618 & \textcolor{red}{0.313} $\pm$ 0.00882 & 0.351 $\pm$ 0.00715 \\
                    &                 & ROUGE-L ($\uparrow$) & 0.113 $\pm$ 0.00534 & \textcolor{red}{0.271} $\pm$ 0.00805 & 0.294 $\pm$ 0.00673 \\
                    \cmidrule{2-6}
                    & Zero-Shot CoT   & ROUGE-1 ($\uparrow$) & \textcolor{blue}{\textbf{0.136}} $\pm$ 0.00655 & \textcolor{blue}{\textbf{0.434}} $\pm$ 0.00962 & \textcolor{red}{0.310} $\pm$ 0.00596  \\
                    &                 & ROUGE-L ($\uparrow$) & 0.120 $\pm$ 0.00574  & \textcolor{blue}{\textbf{0.370}} $\pm$ 0.00928  & \textcolor{red}{0.252} $\pm$ 0.00556 \\
                    \cmidrule{2-6}
                    & Few-Shot CoT    & ROUGE-1 ($\uparrow$) & 0.137 $\pm$ 0.00639 & 0.423 $\pm$ 0.01053 & \textcolor{blue}{\textbf{0.439}} $\pm$ 0.00705 \\
                    &                 & ROUGE-L ($\uparrow$) & 0.122 $\pm$ 0.00565 & 0.368 $\pm$ 0.01033 & \textcolor{blue}{\textbf{0.381}} $\pm$ 0.00681 \\
    \midrule
    \midrule
    \multirow{14}{*}{Retrieved History} & Plain           & ROUGE-1 ($\uparrow$) & 0.138 $\pm$ 0.00614 & 0.419 $\pm$ 0.00953 & 0.367 $\pm$ 0.00648 \\
                    &                 & ROUGE-L ($\uparrow$) & 0.123 $\pm$ 0.00551 & 0.352 $\pm$ 0.00899 & 0.310 $\pm$ 0.00609  \\
                    \cmidrule{2-6}
                    & PreJust         & ROUGE-1 ($\uparrow$) & 0.138 $\pm$ 0.00636 & 0.420 $\pm$ 0.00957  & 0.358 $\pm$ 0.00659 \\
                    &                 & ROUGE-L ($\uparrow$) & 0.123 $\pm$ 0.00570  & 0.356 $\pm$ 0.00932 & 0.306 $\pm$ 0.00619 \\
                    \cmidrule{2-6}
                    & PostJust        & ROUGE-1 ($\uparrow$) & \textcolor{blue}{\textbf{0.143}} $\pm$ 0.00665 & \textcolor{blue}{\textbf{0.443}} $\pm$ 0.01028 & 0.375 $\pm$ 0.00706 \\
                    &                 & ROUGE-L ($\uparrow$) & \textcolor{blue}{\textbf{0.130}} $\pm$ 0.00622  & \textcolor{blue}{\textbf{0.383}} $\pm$ 0.01001 & 0.321 $\pm$ 0.00659 \\
                    \cmidrule{2-6}
                    & CrossDomainJust & ROUGE-1 ($\uparrow$) & \textcolor{red}{0.136} $\pm$ 0.00629 & 0.402 $\pm$ 0.01025 & \textcolor{red}{0.325} $\pm$ 0.00647 \\
                    &                 & ROUGE-L ($\uparrow$) & 0.122 $\pm$ 0.00556 & 0.338 $\pm$ 0.00950  & \textcolor{red}{0.274} $\pm$ 0.00615 \\
                    \cmidrule{2-6}
                    & FakeJust        & ROUGE-1 ($\uparrow$) & 0.138 $\pm$ 0.00644 & \textcolor{red}{0.266} $\pm$ 0.01036 & \textcolor{blue}{\textbf{0.383}} $\pm$ 0.00727 \\
                    &                 & ROUGE-L ($\uparrow$) & 0.126 $\pm$ 0.00594 & \textcolor{red}{0.239} $\pm$ 0.00957 & \textcolor{blue}{\textbf{0.325}} $\pm$ 0.00678 \\
                    \cmidrule{2-6}
                    & Zero-Shot CoT   & ROUGE-1 ($\uparrow$) & 0.137 $\pm$ 0.00634 & 0.427 $\pm$ 0.01004 & 0.349 $\pm$ 0.00657 \\
                    &                 & ROUGE-L ($\uparrow$) & \textcolor{red}{0.121} $\pm$ 0.00566 & 0.363 $\pm$ 0.00980  & 0.293 $\pm$ 0.00604 \\
                    \cmidrule{2-6}
                    & NumericalJust   & ROUGE-1 ($\uparrow$) & \textcolor{blue}{\textbf{0.143}} $\pm$ 0.00657 & 0.425 $\pm$ 0.01066 & 0.377 $\pm$ 0.00721 \\
                    &                 & ROUGE-L ($\uparrow$) & 0.128 $\pm$ 0.00586 & 0.372 $\pm$ 0.01027 & 0.322 $\pm$ 0.00695 \\
    \midrule
    \midrule
    \multirow{4}{*}{Other}           & Entire History  & ROUGE-1 ($\uparrow$) & \textcolor{blue}{\textbf{0.144}} $\pm$ 0.00686 & \textcolor{red}{0.235} $\pm$ 0.01090  & \textcolor{blue}{\textbf{0.433}} $\pm$ 0.00755 \\
                    &                 & ROUGE-L ($\uparrow$) & 0.128 $\pm$ 0.00619 & \textcolor{red}{0.200} $\pm$ 0.00946   & \textcolor{blue}{\textbf{0.379}} $\pm$ 0.00730  \\
                    \cmidrule{2-6}
                    & Random History  & ROUGE-1 ($\uparrow$) & 0.134 $\pm$ 0.00612 & 0.407 $\pm$ 0.00890  & 0.320 $\pm$ 0.00621  \\
                    &                 & ROUGE-L ($\uparrow$) & 0.119 $\pm$ 0.00544 & 0.333 $\pm$ 0.00819 & 0.266 $\pm$ 0.00567\\
    \midrule
    \midrule
\end{tabular}
\label{table:full_results_gpt3}
\end{table*}

\begin{table*}
   \caption{The mean ROUGE-1 and ROUGE-L scores for each method, tested on GPT-4, with accompanying one-sigma standard error bars. The highest ROUGE-1 and ROUGE-L scores in each subsection (No History and Retrieved History) for each dataset are highlighted in blue, while the lowest scores are highlighted in red. The "Other" section (Entire and Random History) is highlighted if the scores are higher than in all the other sections.}
   \centering
\begin{tabular}{|c|c|c|ccc|}
    \hline
    \multirow{2}{*}{} & \multirow{2}{*}{} & \multirow{2}{*}{} & \multicolumn{3}{c}{GPT-4} \\ \midrule 
    Type & Method & Metric & LaMP-4 & LaMP-5 & LaMP-7 \\ \midrule
    \multirow{14}{*}{No History}        & Plain           & ROUGE-1 ($\uparrow$) & \textcolor{red}{0.123} $\pm$ 0.00847 & 0.389 $\pm $0.01703 & 0.268 $\pm $0.01172 \\
         &                 & ROUGE-L ($\uparrow$) & 0.110 $\pm$ 0.00771  & 0.313 $\pm $0.01493 & 0.224 $\pm $0.01062 \\
         \cmidrule{2-6}
         & PreJust         & ROUGE-1 ($\uparrow$) & 0.128 $\pm$ 0.00834 & \textcolor{red}{0.358} $\pm $0.01648 & \textcolor{red}{0.242} $\pm $0.01279 \\
         &                 & ROUGE-L ($\uparrow$) & 0.116 $\pm$ 0.00734 & 0.301 $\pm $0.01539 & \textcolor{red}{0.200} $\pm $0.01125   \\
         \cmidrule{2-6}
         & PostJust        & ROUGE-1 ($\uparrow$) & 0.129 $\pm$ 0.00845 & 0.395 $\pm $0.01802 & 0.270 $\pm $0.01184  \\
         &                 & ROUGE-L ($\uparrow$) & 0.115 $\pm$ 0.00741 & 0.319 $\pm $0.01561 & 0.222 $\pm $0.01029 \\
         \cmidrule{2-6}
         & CrossDomainJust & ROUGE-1 ($\uparrow$) & 0.130 $\pm$ 0.00832  & 0.371 $\pm $0.01672 & 0.285 $\pm $0.01288 \\
         &                 & ROUGE-L ($\uparrow$) & 0.114 $\pm$ 0.00732 & \textcolor{red}{0.299} $\pm $0.01500   & 0.240 $\pm $0.01208  \\
         \cmidrule{2-6}
         & FakeJust        & ROUGE-1 ($\uparrow$) & 0.131 $\pm$ 0.00919 & 0.387 $\pm $0.01819 & 0.260 $\pm $0.01205  \\
         &                 & ROUGE-L ($\uparrow$) & 0.115 $\pm$ 0.00808 & 0.311 $\pm $0.01645 & 0.210 $\pm $0.00994  \\
         \cmidrule{2-6}
         & Zero-Shot CoT   & ROUGE-1 ($\uparrow$) & 0.125 $\pm$ 0.00836 & 0.400 $\pm $0.01783   & 0.268 $\pm $0.01217 \\
         &                 & ROUGE-L ($\uparrow$) & \textcolor{red}{0.109} $\pm$ 0.00724 & \textcolor{blue}{\textbf{0.331}} $\pm $0.01660  & 0.215 $\pm $0.00964 \\
         \cmidrule{2-6}
         & Few-Shot CoT    & ROUGE-1 ($\uparrow$) & \textcolor{blue}{\textbf{0.135}} $\pm$ 0.00947 & \textcolor{blue}{\textbf{0.401}} $\pm $0.01819 & \textcolor{blue}{\textbf{0.341}} $\pm $0.01414 \\
         &                 & ROUGE-L ($\uparrow$) & \textcolor{blue}{\textbf{0.117}} $\pm$ 0.00827 & 0.321 $\pm $0.01610  & \textcolor{blue}{\textbf{0.294}} $\pm $0.01330  \\
         \midrule
         \midrule
    \multirow{14}{*}{Retrieved History} & Plain           & ROUGE-1 ($\uparrow$) & 0.139 $\pm$ 0.0092  & 0.400 $\pm $0.01769   & 0.283 $\pm $0.01189 \\
         &                 & ROUGE-L ($\uparrow$) & 0.126 $\pm$ 0.00863 & \textcolor{red}{0.319} $\pm $0.01642 & 0.232 $\pm $0.01031 \\
         \cmidrule{2-6}
         & PreJust         & ROUGE-1 ($\uparrow$) & 0.141 $\pm$ 0.00957 & \textcolor{red}{0.397} $\pm $0.01881 & 0.279 $\pm $0.01335 \\
         &                 & ROUGE-L ($\uparrow$) & 0.129 $\pm$ 0.00893 & 0.336 $\pm $0.01808 & 0.229 $\pm $0.01188 \\
         \cmidrule{2-6}
         & PostJust        & ROUGE-1 ($\uparrow$) & \textcolor{blue}{\textbf{0.150}} $\pm$ 0.01000     & \textcolor{blue}{\textbf{0.438}} $\pm $0.01946 & 0.273 $\pm $0.01245 \\
         &                 & ROUGE-L ($\uparrow$) & \textcolor{blue}{\textbf{0.138}} $\pm$ 0.00941 & \textcolor{blue}{\textbf{0.351}} $\pm $0.01727 & 0.226 $\pm $0.01109 \\
         \cmidrule{2-6}
         & CrossDomainJust & ROUGE-1 ($\uparrow$) & 0.144 $\pm$ 0.00905 & 0.400 $\pm $0.01838   & \textcolor{red}{0.177} $\pm $0.01636 \\
         &                 & ROUGE-L ($\uparrow$) & 0.129 $\pm$ 0.00791 & 0.325 $\pm $0.01593 & \textcolor{red}{0.142} $\pm $0.01367 \\
         \cmidrule{2-6}
         & FakeJust        & ROUGE-1 ($\uparrow$) & 0.139 $\pm$ 0.00953 & 0.401 $\pm $0.01893 & 0.277 $\pm $0.01155 \\
         &                 & ROUGE-L ($\uparrow$) & 0.125 $\pm$ 0.00856 & 0.332 $\pm $0.01743 & 0.232 $\pm $0.01077 \\
         \cmidrule{2-6}
         & Zero-Shot CoT   & ROUGE-1 ($\uparrow$) & \textcolor{red}{0.135} $\pm$ 0.00890  & 0.402 $\pm $0.01757 & 0.272 $\pm $0.01276 \\
         &                 & ROUGE-L ($\uparrow$) & \textcolor{red}{0.122} $\pm$ 0.00823 & 0.331 $\pm $0.01666 & 0.226 $\pm $0.01132 \\
         \cmidrule{2-6}
         & NumericalJust   & ROUGE-1 ($\uparrow$) & 0.143 $\pm$ 0.01001 & 0.426 $\pm $0.01874 & \textcolor{blue}{\textbf{0.290}} $\pm $0.01285  \\
         &                 & ROUGE-L ($\uparrow$) & 0.130 $\pm$ 0.00929  & 0.343 $\pm $0.01708 & \textcolor{blue}{\textbf{0.239}} $\pm $0.01147 \\
         \midrule
         \midrule
    \multirow{4}{*}{Other}              & Entire History  & ROUGE-1 ($\uparrow$) & 0.146 $\pm$ 0.00948 & 0.381 $\pm $0.01707 & 0.319 $\pm $0.01534 \\
         &                 & ROUGE-L ($\uparrow$) & 0.133 $\pm$ 0.00877 & 0.311 $\pm $0.01574 & 0.280 $\pm $0.01501  \\
         \cmidrule{2-6}
         & Random History  & ROUGE-1 ($\uparrow$) & \textcolor{red}{0.122} $\pm$ 0.00851 & 0.383 $\pm $0.01750  & 0.260 $\pm $0.01132  \\
         &                 & ROUGE-L ($\uparrow$) & 0.110 $\pm$ 0.00790   & 0.303 $\pm $0.01509 & 0.217 $\pm $0.01039\\
    \midrule
    \midrule
\end{tabular}
\label{table:full_results_gpt4}
\end{table*}

\subsection{Participant Demographics}

\begin{figure}
    \centering
    \includegraphics[width=0.5\textwidth]{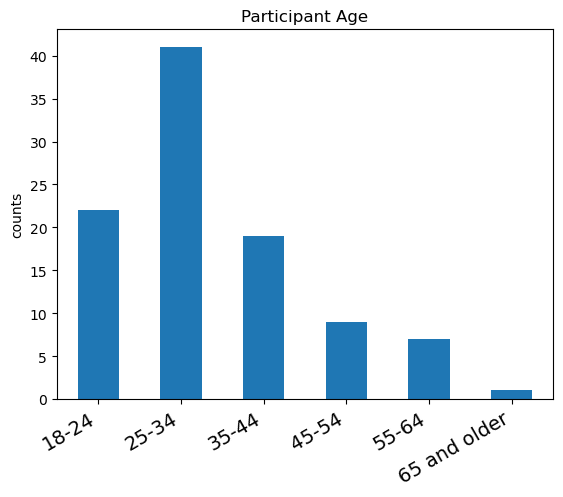}
    \caption{Participant age distribution.}
    \label{fig:demo-age}
\end{figure}

\begin{figure}
    \centering
    \includegraphics[width=0.5\textwidth]{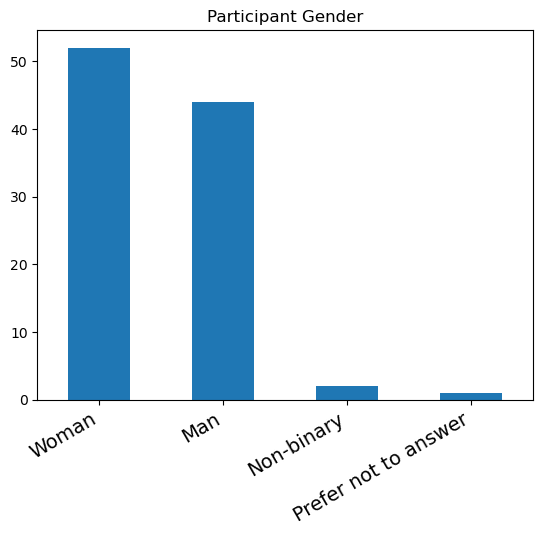}
    \caption{Participant gender distribution. N=3 for groups other than male/female, so we did not perform statistical tests for those groups.}
    \label{fig:demo-gender}
\end{figure}

\begin{figure}
    \centering
    \includegraphics[width=0.5\textwidth]{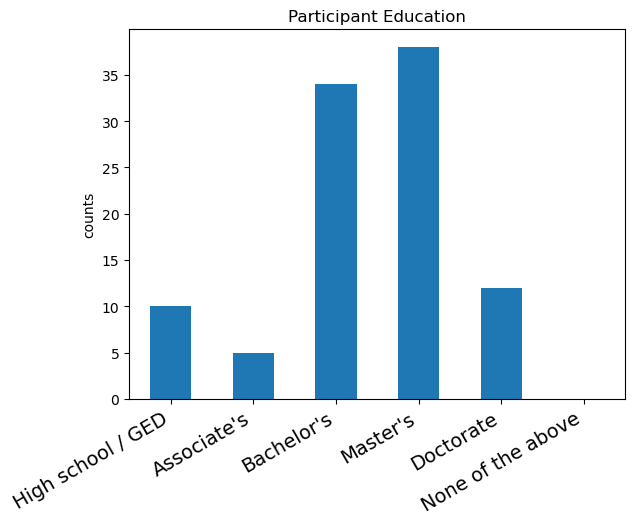}
    \caption{Participant education distribution.}
    \label{fig:demo-education}
\end{figure}

\begin{figure}
    \centering
    \includegraphics[width=0.5\textwidth]{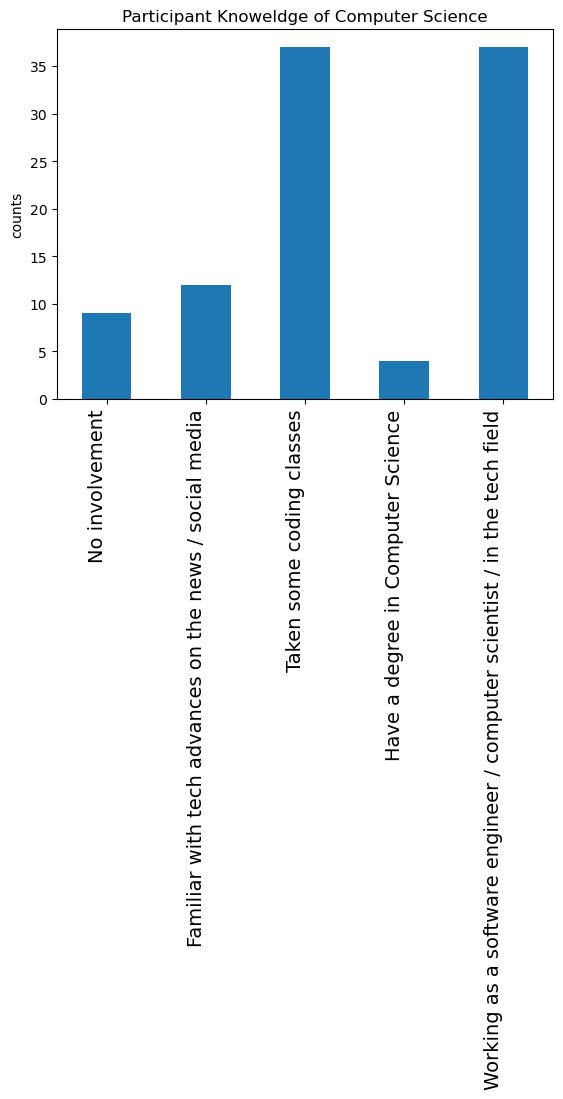}
    \caption{Participant computer science experience distribution.}
    \label{fig:demo-cs}
\end{figure}

\begin{figure}
    \centering
    \includegraphics[width=0.5\textwidth]{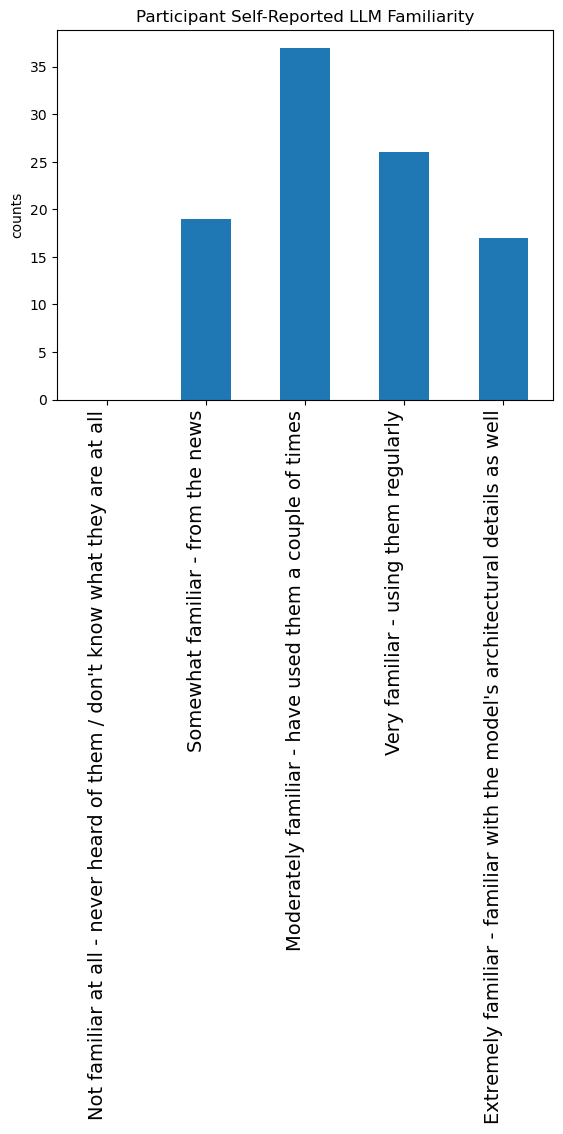}
    \caption{Participant experience with LLMs.}
    \label{fig:demo-llm}
\end{figure}

\begin{figure}
    \centering
    \includegraphics[width=0.5\textwidth]{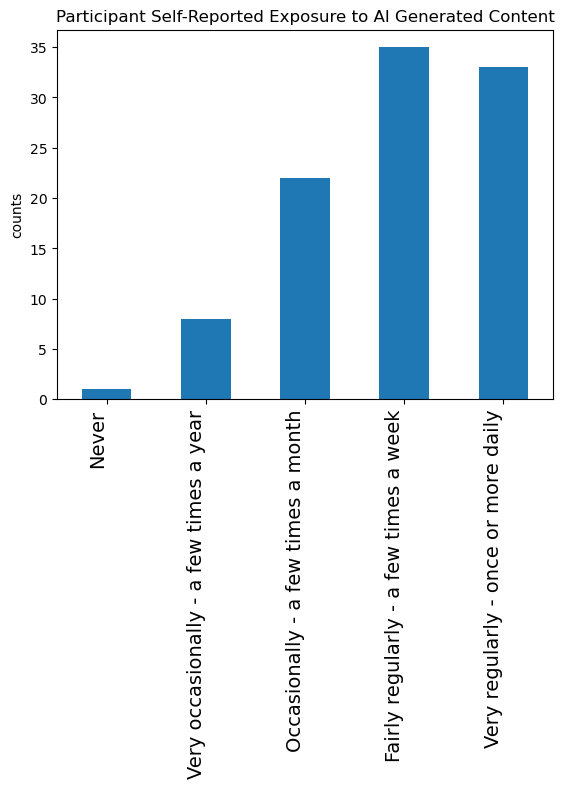}
    \caption{Participant self-reported experience with AI-generated content.}
    \label{fig:demo-ai}
\end{figure}

Additional plots of participant demographic distributions are show in Figures \ref{fig:demo-age}, \ref{fig:demo-gender}, \ref{fig:demo-education}, \ref{fig:demo-cs}, \ref{fig:demo-llm}, and \ref{fig:demo-ai}.

\subsection{Additional Limitations}
Even though self-explanations may not accurately reflect the model’s actual reasoning process \cite{jacovi-goldberg-2020-towards}, there is some evidence they track with traditional methods like occlusion and LIME \cite{huang2023large}. Additionally, as Doshi-Velez and Kim (2017) argue, a true estimate of faithfulness is ultimately impossible for models that are not interpretable per se, which includes LLMs \cite{doshivelez2017rigorous}.

The ROUGE score is also not a perfect metric for this application - for open-ended and creative tasks such as news headline generation, there are many correct answers that could be given (not just the ground truth headline that was suggested in reality) and so proximity to a ground-truth metric is not the only way to measure model performance. One other metric is similarity to the question asked and information provided (in this case, the article). An even better evaluation is using human feedback, which we acquired in this experiment. On the whole, however, we believe that the ML domain requires better metrics for open-ended responses on the model side.

\end{document}